\documentclass[10pt,twocolumn,letterpaper]{article}

\usepackage[pagenumbers]{cvpr} 

\usepackage[accsupp]{axessibility} 

\usepackage{epsfig}
\usepackage{dsfont}
\usepackage{graphicx}
\usepackage{amsmath}
\usepackage{amssymb}
\usepackage{booktabs}
\usepackage{color}
\usepackage{cite,eucal}
\usepackage{multirow,xcolor}

\definecolor{citecolor}{HTML}{0071bc}
\usepackage[pagebackref,breaklinks=true,letterpaper=true,colorlinks,citecolor=citecolor,bookmarks=false]{hyperref}

\usepackage[capitalize]{cleveref}
\crefname{section}{Sec.}{Secs.}
\Crefname{section}{Section}{Sections}
\Crefname{table}{Table}{Tables}
\crefname{table}{Tab.}{Tabs.}

\DeclareMathOperator*{\argmin}{arg\,min}

\newcommand{\xvec}{{\mathbf{x}}}
\newcommand{\yxvec}{{y_{\mathbf{x}}}}

\def\cvprPaperID{6784} 
\def\confName{CVPR}
\def\confYear{2022}

\begin{document}


\title{Catching Both Gray and Black Swans: Open-set Supervised Anomaly Detection\thanks{Corresponding author: CS (e-mail: {\tt  chunhua@me.com}). This work was in part
done when GP and CS were with The University of Adelaide.}
}

\author{Choubo Ding$^{1\dagger}$,\qquad Guansong Pang$^2$\thanks{First two authors contributed equally.},\qquad Chunhua Shen$^3$\\[0.12cm]
        $^1$ The University of Adelaide\quad $^2$ Singapore Management University\quad $^3$ Zhejiang University}

\maketitle

\begin{abstract}
Despite most 
existing anomaly detection studies assume the availability of normal training samples only, a few labeled anomaly examples are often available in many real-world applications, such as defect samples identified during random quality inspection, lesion images confirmed by radiologists in daily medical screening, etc. These anomaly examples provide valuable knowledge about the application-specific abnormality, enabling significantly improved detection of similar anomalies in some recent models. However, those anomalies seen during training often do not illustrate every possible class of anomaly, rendering these models ineffective in generalizing to unseen anomaly classes.
This paper tackles open-set supervised anomaly detection, in which we learn detection models using the anomaly examples with the objective to detect both seen anomalies (`gray swans') and unseen anomalies (`black swans'). We propose a novel approach that learns disentangled representations of abnormalities illustrated by seen anomalies, pseudo anomalies, and latent residual anomalies (i.e., samples that have unusual residuals compared to the normal data in a latent space), with the last two abnormalities designed to detect unseen anomalies. Extensive experiments on nine real-world anomaly detection datasets show superior performance of our model in detecting seen and unseen anomalies
under diverse settings. Code and data are available at: \renewcommand\UrlFont{\color{blue}\tt}
\url{https://github.com/choubo/DRA}

\end{abstract}


\section{Introduction}
\begin{figure}[t] 
    \centering
    \includegraphics[width=0.85\linewidth]{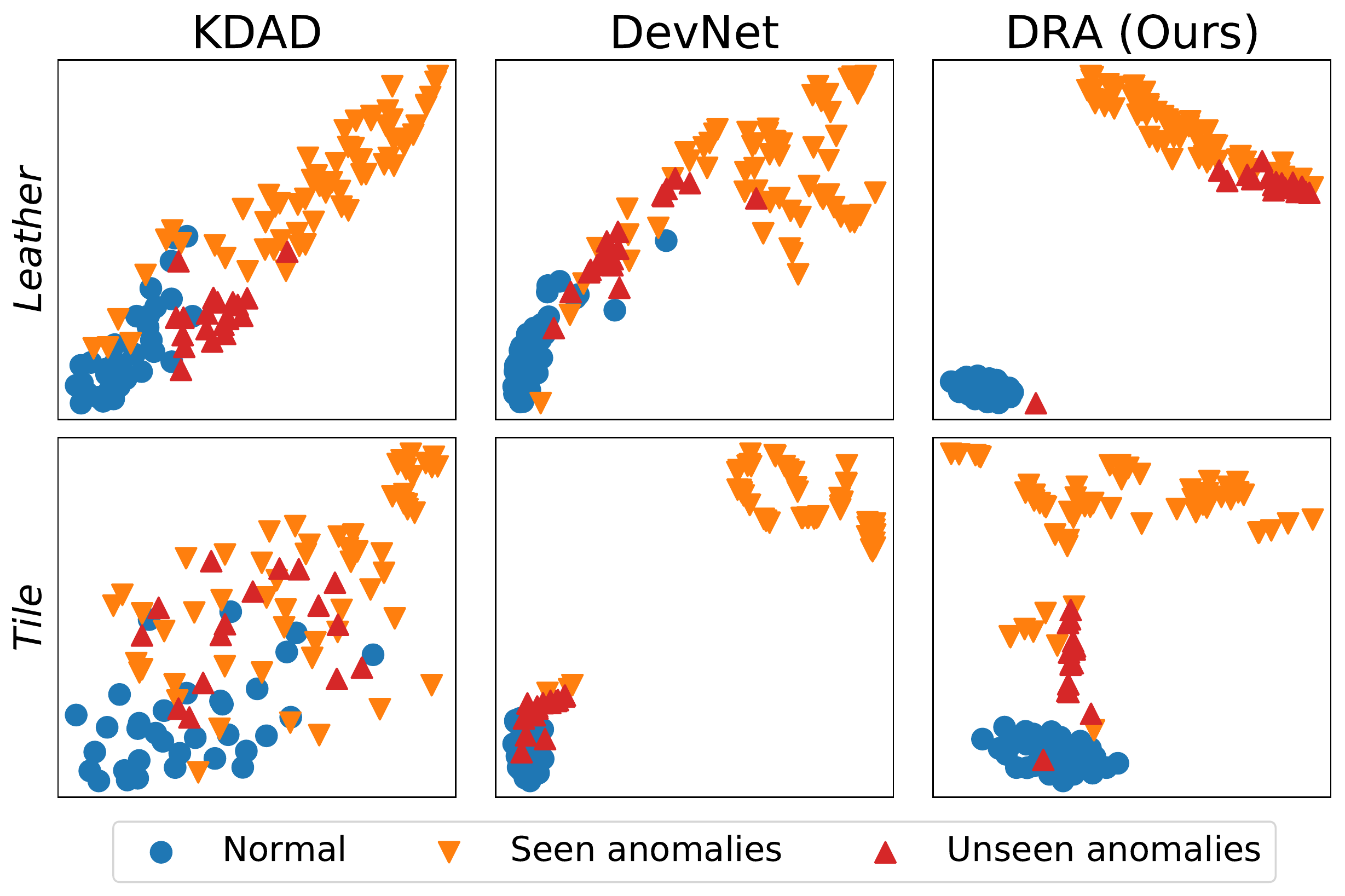}
    \vspace{-0.2cm}
    \caption{t-SNE visualization of features learned by SotA unsupervised (KDAD \cite{salehi2021multiresolution}) and  supervised (DevNet \cite{pang2019deep,pang2021explainable}) models, and our open-set supervised model (DRA) on the test data of two MVTec AD datasets, Leather and Tile. KDAD is trained with normal data only, learning less discriminative features than DevNet and DRA that are trained using ten samples from the seen anomaly classes, in addition to the normal data. DevNet is prone to overfitting the seen anomalies, failing to distinguish unseen anomalies from the normal data, while DRA effectively mitigates this issue.}
    \label{fig:tsne}
    \vspace{-0.5cm}
\end{figure}

Anomaly detection (AD) 
aims at identifying exceptional samples that do not conform to expected patterns \cite{pang2021deep}. It has broad applications in diverse domains, \eg, lesion detection in medical image analysis \cite{schlegl2019f,zhang2020viral,tian2021constrained}, inspecting micro-cracks/defects in industrial inspection \cite{Bergmann_2019_CVPR,Bergmann_2020_CVPR}, crime/accident detection in video surveillance \cite{tudor2017unmasking,sultani2018real,georgescu2021anomaly,zaheer2020claws}, and unknown object detection in autonomous driving \cite{di2021pixel,tian2021pixel}. Most of existing anomaly detection methods \cite{wang2016s, salehi2021multiresolution, georgescu2021anomaly, wang2021glancing, reiss2021panda, bergman2020goad, Markovitz_2020_CVPR, Park_2020_CVPR, venkataramanan2020attention, zhou2020encoding,gong2019memorizing, Park_2020_CVPR, schlegl2019f, sabokrou2018adversarially, zaheer2020old, ruff2018deep,chen2021deep}
are unsupervised, which assume the availability of normal training samples only, \ie, anomaly-free training data, because it is difficult, if not impossible, to collect large-scale anomaly data. However, a small number of (\eg, one to multiple) labeled anomaly examples are often available in many relevant real-world applications, such as some defect samples identified during random quality inspection, lesion images confirmed by radiologists in daily medical screening, etc. These anomaly examples provide valuable knowledge about application-specific abnormality \cite{ruff2019deep,liu2019margin, pang2019deep, pang2021explainable}, but the unsupervised detectors are unable to utilize them. Due to the lack of knowledge about anomalies, the learned features in unsupervised models are not discriminative enough to distinguish anomalies (especially some challenging ones) from normal data, as illustrated by the results of KDAD \cite{salehi2021multiresolution}, a recent state-of-the-art (SotA) unsupervised method, on two MVTec AD defect detection datasets \cite{Bergmann_2019_CVPR} in Fig. \ref{fig:tsne}.

In recent years, there have been some studies \cite{ruff2019deep,pang2019deep,pang2021explainable,liu2019margin} exploring a supervised detection paradigm that  
aims at exploiting those small, readily accessible anomaly data---rare but previously occurred exceptional cases/events, \textit{a.k.a.}\ \texttt{gray swans} \cite{khakzad2015grayswan} -- to train anomaly-informed detection models. The current methods in this line focus on fitting these anomaly examples using one-class metric learning with the anomalies as negative samples \cite{ruff2019deep,liu2019margin} or one-sided anomaly-focused deviation loss \cite{pang2019deep,pang2021explainable}. Despite the limited amount of the anomaly data, they achieve largely improved performance in detecting anomalies that are similar to the anomaly examples seen during training. However, these seen anomalies often do not illustrate every possible class of anomaly because i) anomalies per se are unknown and ii) the seen and unseen anomaly classes can differ largely from each other \cite{pang2021deep}, \eg, the defective features of color stains are very different from that of folds and cuts in leather defect inspection. Consequently, these models can overfit the seen anomalies,
failing to generalize to unseen/unknown anomaly classes---rare and previously unknown exceptional cases/events, \textit{a.k.a.}\  \texttt{black swans} \cite{taleb2007black}, as shown by the result of DevNet \cite{pang2019deep,pang2021explainable} in Fig. \ref{fig:tsne} where DevNet improves over KDAD in detecting the seen anomalies but fails to discriminate unseen anomalies from normal samples. In fact, these supervised models can be biased by the given anomaly examples and become less effective in detecting unseen anomalies than unsupervised detectors (see DevNet vs.\  KDAD on the Tile dataset in Fig. \ref{fig:tsne}).

To address this issue, this paper tackles open-set supervised anomaly detection, in which detection models are trained using the small anomaly examples in an open-set environment, \ie, the objective is to detect both seen anomalies (`gray swans') and unseen anomalies (`black swans').
To this end, we propose a novel anomaly detection approach, termed DRA, that learns \textbf{\underline{d}isentangled \underline{r}epresentations of \underline{a}bnormalities} to enable the generalized detection.
Particularly, we disentangle the unbounded abnormalities into three general categories: anomalies similar to the limited seen anomalies, anomalies that are similar to pseudo anomalies created from data augmentation or external data sources, and unseen anomalies that are detectable in some latent residual-based composite feature spaces. We further devise a multi-head network, with separate heads enforced to learn each type of these three disentangled abnormalities. In doing so, our model learns diversified abnormality representations rather than only the known abnormality, which can discriminate both seen and unseen anomalies from the normal data, as shown in Fig. \ref{fig:tsne}.

In summary, we make the following main contributions:
\begin{itemize}
\itemsep -0.1cm 
    \item To tackle open-set supervised AD, we propose to learn disentangled representations of abnormalities illustrated by seen anomalies, pseudo anomalies, and latent residual-based anomalies. This learns diversified abnormality representations, extending the set of anomalies sought to both seen and unseen anomalies.
    \item We propose a novel multi-head neural network-based model DRA to learn the disentangled abnormality representations, with each head dedicated to capturing one specific type of abnormality. 
    \item We further introduce a latent residual-based abnormality learning module that learns abnormality upon the residuals between the intermediate feature maps of normal and abnormal samples. This helps learn discriminative composite features for the detection of hard anomalies (\eg, unseen anomalies) that cannot be detected in the original non-composite feature space.
    \item We perform comprehensive experiments on nine real-application datasets from industrial inspection, rover-based planetary exploration and medical image analysis. The results show that our model substantially outperforms five SotA competing models in diverse settings. The results also establish new baselines for future work in this important emerging direction. 
\end{itemize}
    
\begin{figure*}[t]
\centering     
\subcaptionbox{\label{fig:a}}{\includegraphics[height=48mm]{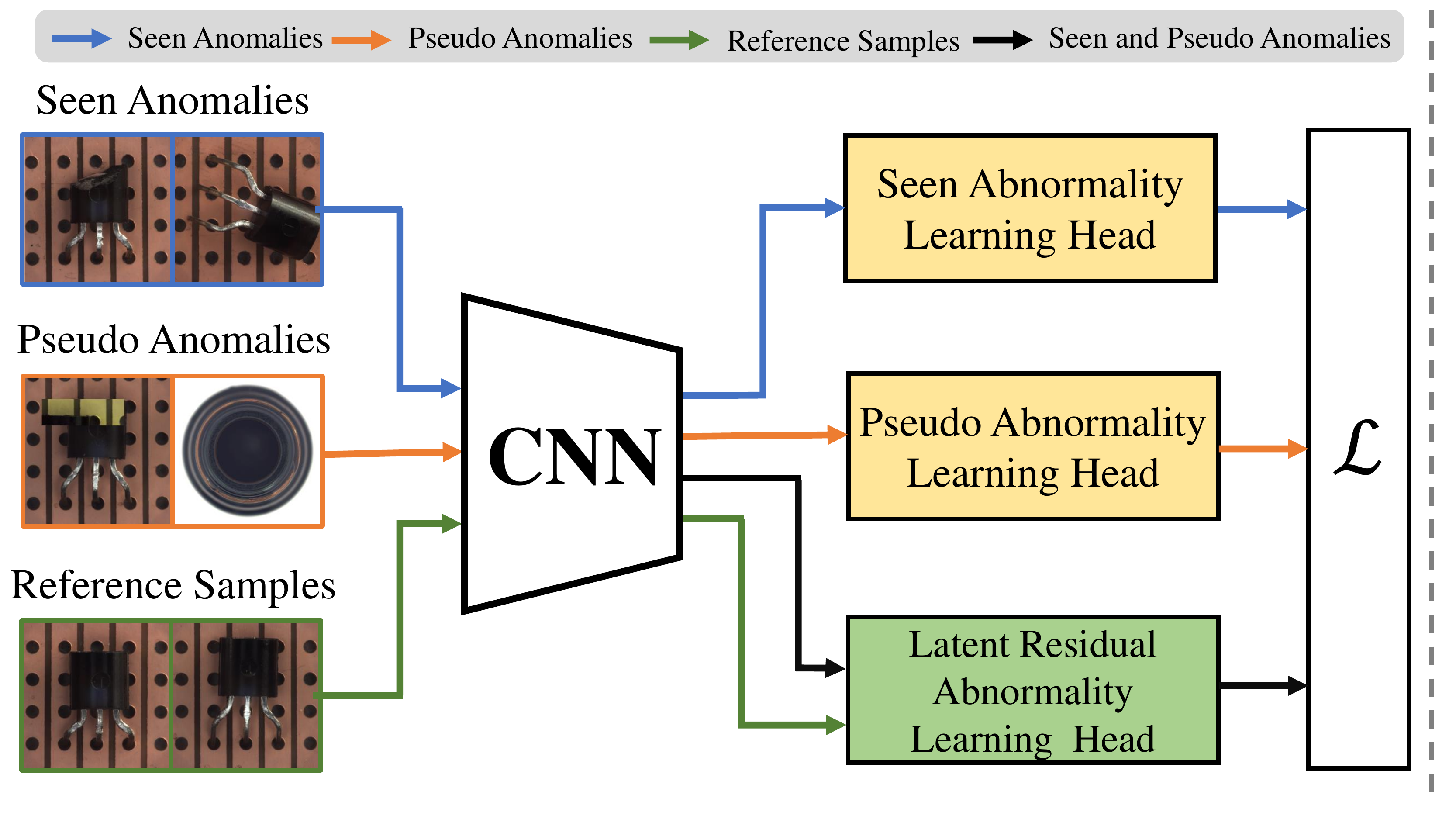}}
\hspace{1pt}
\subcaptionbox{\label{fig:b}}{\includegraphics[height=48mm]{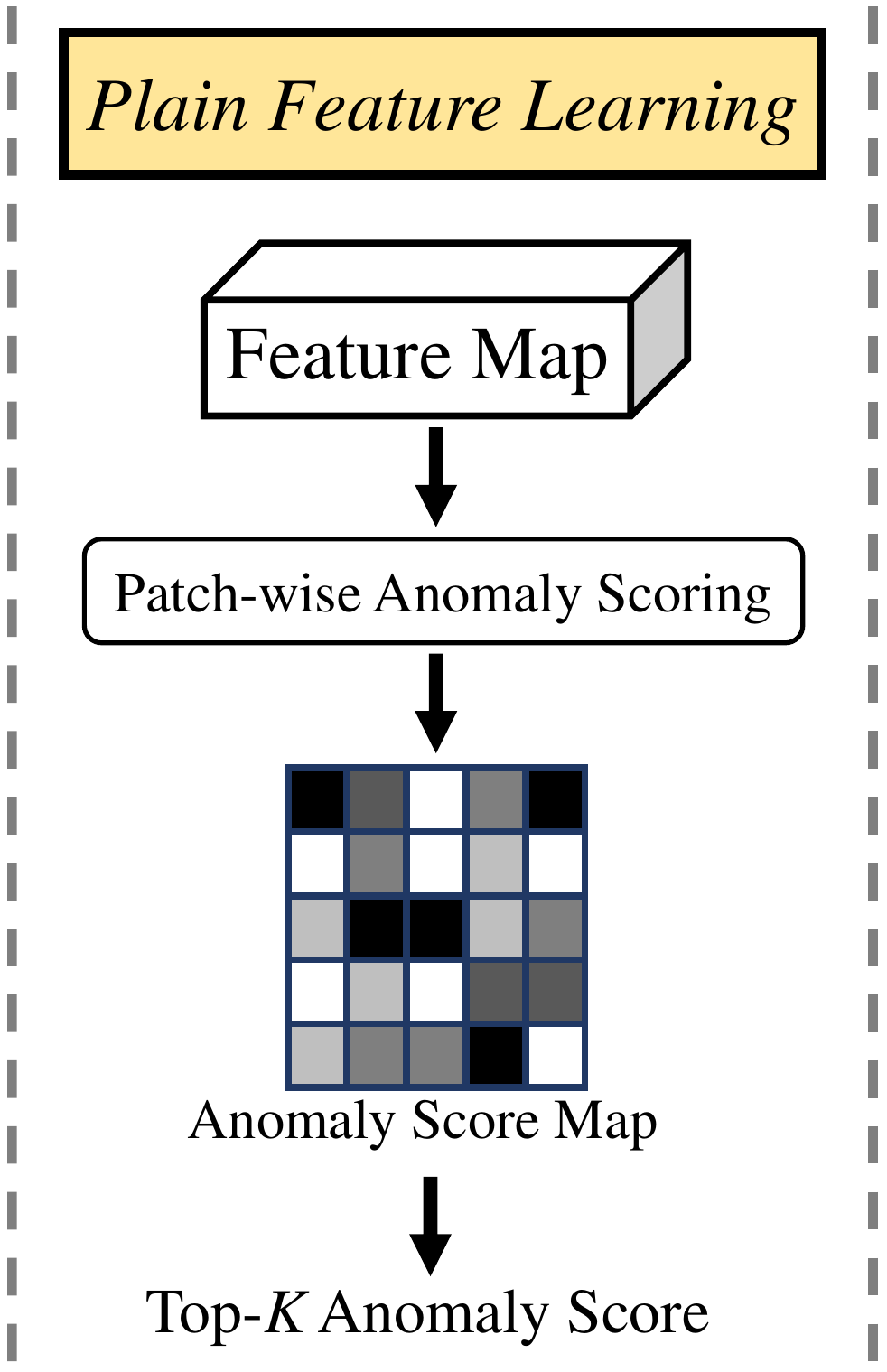}}
\vspace{1pt}
\subcaptionbox{\label{fig:c}}{\includegraphics[height=48mm]{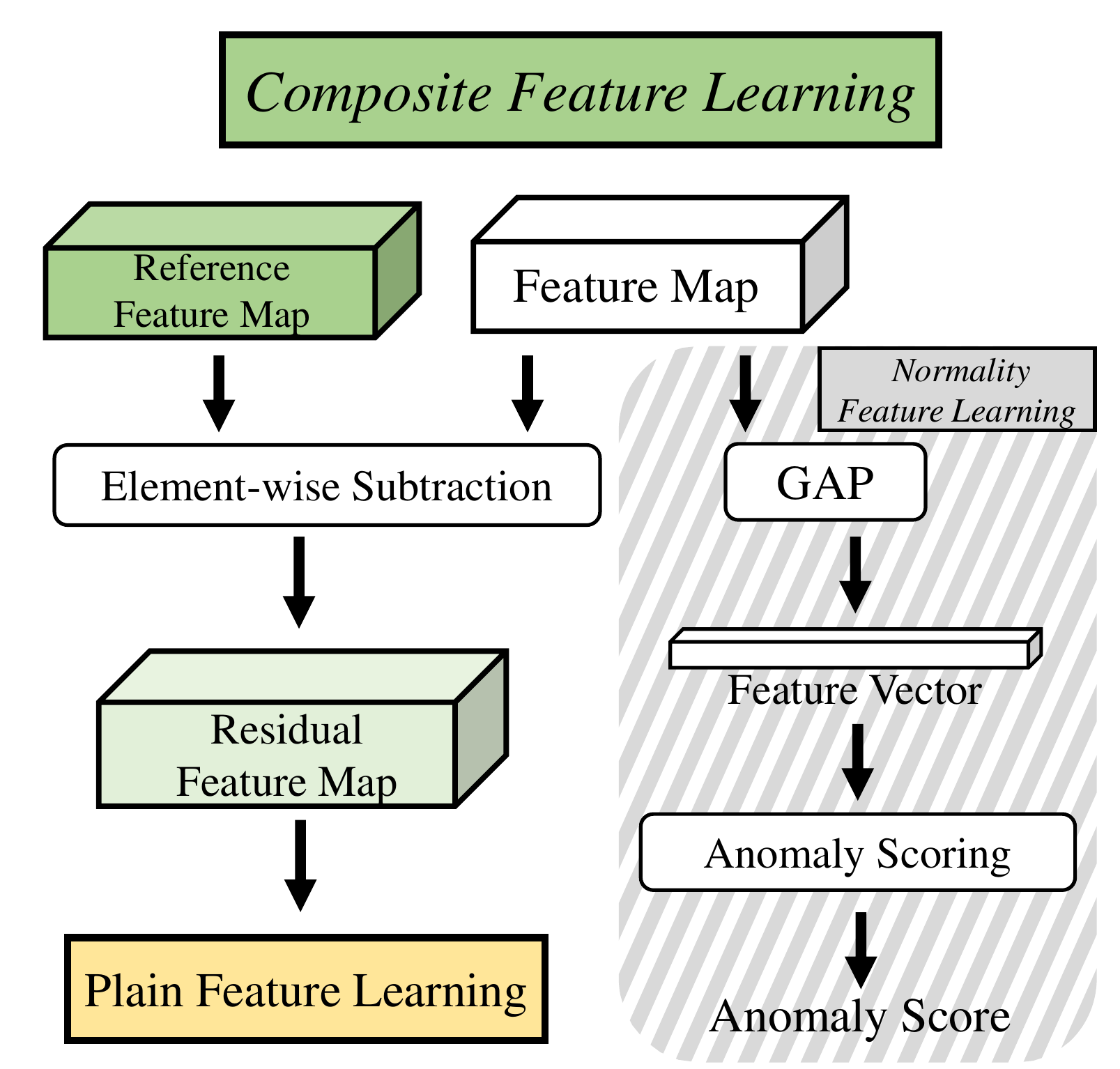}}
\vspace{1pt}
    \vspace{-0.3cm}
\caption{Overview of our proposed framework. (a) presents the high-level procedure of learning three disentangled abnormalities, (b) shows the abnormality feature learning in the plain (non-composite) feature space for the seen and pseudo abnormality learning heads, and (c) shows the framework of our proposed latent residual abnormality learning in a composite feature space.
}
    \vspace{-0.4cm}
\end{figure*}

\section{Related Work}
\textbf{Unsupervised Approaches}. Most existing anomaly detection methods, such as autoencoder-base methods \cite{zhou2017anomaly, gong2019memorizing, Hou_2021_ICCV, Park_2020_CVPR, zhou2020encoding}, GAN-base methods \cite{schlegl2019f, Perera_2019_CVPR, sabokrou2018adversarially, zaheer2020old}, self-supervised methods \cite{golan2018deep,wang2019effective,bergman2020goad,georgescu2021anomaly, sohn2020learning, li2021cutpaste,tian2021constrained}, and one-class classification methods \cite{perera2019learning, ruff2018deep, chalapathy2018anomaly,chen2021deep}, assume that only normal data can be accessed during training.
Although they do not have the risk of biasing towards the seen anomalies, they are difficult to distinguish anomalies from normal samples due to the lack of knowledge about true anomalies.

\textbf{Supervised Approaches}. A recently emerging direction focuses on supervised (or semi-supervised) anomaly detection that alleviates the lack of anomaly information by leveraging small anomaly examples to learn anomaly-informed models. This is achieved by one-class metric learning with the anomalies as negative samples \cite{pang2018learning,ruff2019deep,liu2019margin,gornitz2013toward} or one-sided anomaly-focused deviation loss \cite{pang2019deep,pang2021explainable,zhang2020viral}. However, these models rely heavily on the seen anomalies and can overfit the known abnormality. A reinforcement learning approach is introduced in \cite{pang2021toward} to mitigate this overfitting issue, but it assumes the availability of large-scale unlabeled data and the presence of unseen anomalies in those data.
Supervised anomaly detection is similar to imbalanced classification \cite{he2009imbalance, branco2016survey, liu2019large} in that they both detect rare classes with
a few labeled examples. However, 
due to the unbound nature and unknowingness of anomalies, anomaly detection is inherently an open-set task, while the imbalanced classification task is typically formulated as a closed-set problem. 

\textbf{Learning In- and Out-of-distribution}. Out-of-distribution (OOD) detection \cite{hendrycks17baseline, ren2019likelihood,huang2021mos, Lin_2021_CVPR, Zaeemzadeh_2021_CVPR, hendrycks2018deep} and open-set recognition \cite{scheirer2012toward,bendale2016towards, zhou2021learning, Yue_2021_CVPR, liu2019margin} are related tasks to ours. However, they aim at guaranteeing accurate multi-class inlier classification while detecting OOD/uncertain samples, whereas our task is focused on anomaly detection exclusively.
Further, despite the use of pseudo anomalies like outlier exposure \cite{hendrycks2018deep,huang2021mos} shows effective performance, the current models in these two tasks are also assumed to be inaccessible to any true anomalous samples.

\section{Proposed Approach}

\textbf{Problem Statement}
The studied problem, open-set supervised AD, can be formally stated as follows.
Given a set of training samples $\mathcal{X}=\{ \mathbf{x}_{i}
\}_{i=1}^{N+M}$, in which $\mathcal{X}_{n}=\{ \mathbf{x}_{1}, \mathbf{x}_{2}, \cdots, \mathbf{x}_{N}\}$ is the normal sample set and $\mathcal{X}_{a}=\{\mathbf{x}_{N+1}, \mathbf{x}_{N+2}, \cdots, \mathbf{x}_{N+M} \}$ ($M\ll N$) is a very small set of annotated anomalies that provide some knowledge about true anomalies, and the $M$ anomalies belong to the seen anomaly classes $\mathcal{S}\subset\mathcal{C}$, where $\mathcal{C}=\{ c_i\}_{i=1}^{|\mathcal{C}|}$ denotes the set of all possible anomaly classes,
and then the goal is to detect both seen and unseen anomaly classes by learning an anomaly scoring function $g: \mathcal{X} \rightarrow \mathbb{R}$ that assigns larger anomaly scores to both seen and unseen anomalies than normal samples.

\subsection{Overview of Our Approach} \label{sec:4.1}
Our proposed approach DRA is designed
to learn disentangled representations of diverse abnormalities to effectively detect both seen and unseen anomalies.
The learned abnormality representations include the seen abnormality illustrated by the limited given anomaly examples, and the unseen abnormalities illustrated by pseudo anomalies and \textit{latent residual anomalies} (\ie, samples that have unusual residuals compared to normal examples in a learned feature space). In doing so, DRA mitigates the issue of biasing towards seen anomalies and learns generalized detection models. The high-level overview of our proposed framework is provided in Fig. \ref{fig:a}, which is composed of three main modules, including seen, pseudo, and latent residual abnormality learning heads. The first two heads learn abnormality representations in a plain (non-composite) feature space, as shown in Fig. \ref{fig:b}, while the last head learns composite abnormality representations by looking into the deviation of the residual features of input samples to some reference (\ie, normal) images in a learned feature space, as shown in Fig. \ref{fig:c}.
Particularly, given a feature extraction network $f: \mathcal{X} \rightarrow \mathcal{M}$ for extracting the intermediate feature map $\mathbf{M}\in \mathcal{M} \subset \mathbb{R}^{c^\prime\times h^\prime\times w^\prime}$ from a training image $\mathbf{x}\in \mathcal{X} \subset \mathbb{R}^{c\times h\times w}$, and a set of abnormality learning heads $\mathcal{G}=\{g_i\}^{|\mathcal{G}|}_{i=1}$, where each head $g:\mathcal{M}\rightarrow\mathbb{R}$ learns an anomaly score for one type of abnormality, then the overall objective of DRA can be given as follows:

\begin{equation}
\argmin_{\Theta}\sum_{i=1}^{|\mathcal{G}|} \ell_i\big(g_i(f (\mathbf{x};\Theta_f);\Theta_{i}), y_{\xvec}\big),
\end{equation}
where $\Theta$ contains all weight parameters, $y_{\xvec}$ denotes the supervision information of $\xvec$, and $\ell_i$ denotes a loss function for one head. The feature network $f$ is jointly optimized by all the downstream abnormality learning heads, while these heads are independent from each other in learning the specific abnormality. 
Below we introduce each head in detail.
\subsection{Learning Disentangled Abnormalities}

\textbf{Abnormality Learning with Seen Anomalies}.
Most real-world anomalies have only some subtle differences from normal images, sharing most of the common features with normal images.
Patch-wise anomaly learning \cite{yi2020patch, Bergmann_2020_CVPR, wang2021glancing,pang2021explainable} that learns anomaly scores for each small image patch has shown impressive performance in tackling this issue.
Motivated by this, DRA utilizes a top-$K$ multiple-instance-learning (MIL) -based method in \cite{pang2021explainable} to effectively learn the seen abnormality. 
As shown in Fig. \ref{fig:b}, for the feature map $\mathbf{M}_{\xvec}$ of each input image $\xvec$, we generate pixel-wise vector representations $\mathcal{D}=\{\mathbf{d}_i\}_{i=1}^{h^\prime\times w^\prime}$, each of which corresponds to 
the feature vector of a small patch of the input image. These patch-wise representations are then mapped to learn the anomaly scores of the image patches by an anomaly classifier $g_s: \mathcal{D} \rightarrow \mathbb{R}$.
Since only selective image patches contain abnormal features, 
we utilize an optimization using top-$K$ MIL to learn an anomaly score for an image based on the $K$ most anomalous image patches, with the loss function defined as follows:
\begin{equation}\label{eqn:mil}
    \ell_{s}(\mathbf{x}, \yxvec) = \ell\big(g_s(\mathbf{M}_{\xvec}; \Theta_{s}), \yxvec\big),
\end{equation}
where $\ell$ is a binary classification loss function; $\yxvec = 1$ if $\xvec$ is a seen anomaly, and $\yxvec = 0$ if $\xvec$ is a normal sample otherwise; and 
\begin{equation}\label{eqn:seenabnormality}
g_s(\mathbf{M}_{\xvec}; \Theta_{s}) = \max_{\Psi_K(\mathbf{M}_{\xvec})\subset \mathcal{D}} \frac{1}{K} \sum_{\mathbf{d}_{i} \in \Psi_K(\mathbf{M}_{\xvec})} g_s(\mathbf{d}_{i}; \Theta_{s})
\end{equation}
where $\Psi_K(\mathbf{M}_{\xvec})$ is a set of $K$ vectors that have the largest anomaly scores among all vectors in $\mathbf{M}_{\xvec}$. 
\textbf{Abnormality Learning with Pseudo Anomalies}.
We further design a separate head to learn abnormalities that are different from the seen anomalies and simulate some possible classes of unseen anomaly.  
There are two effective methods to create such pseudo anomalies, including data augmentation-based methods \cite{li2021cutpaste,tack2020csi} and outlier exposure \cite{hendrycks2018deep,reiss2021panda}. 
Particularly, for the data augmentation-based method,
we adapt the popular method CutMix \cite{yun2019cutmix} to generate pseudo anomalies $\tilde{\mathbf{x}}$ from normal images $\mathbf{x}_n$ for training, which is defined as follows: 
\begin{equation}\label{eqn:dataaugmentation}
    \tilde{\mathbf{x}}= T\circ C(\mathbf{R}\odot \mathbf{x}_n) + \big(\mathbf{1}-T(\mathbf{R})\big) \odot \mathbf{x}_n
\end{equation}
where $\mathbf{R} \in \{0, 1\}^{h\times w}$ denotes a binary mask of random rectangle, $\mathbf{1}$ is an all-ones matrix, $\odot$ is element-wise multiplication, $T(\cdot)$ is a randomly translate transformation, and $C(\cdot)$ is a random color jitter. 
As shown in Fig. \ref{fig:a}, the pseudo abnormality learning uses the same architecture and anomaly scoring method as the seen abnormality learning to learn fine-grained pseudo abnormal features:
\begin{equation}\label{eqn:pseudoanomaly}
    \ell_{p}(\xvec, y_{\xvec}) = \ell\big(g_p(\mathbf{M}_{\xvec};\Theta_p), y_{\xvec}\big),
\end{equation}
where $\yxvec = 1$ if $\xvec$ is a pseudo anomaly, \ie, $\xvec=\tilde{\xvec}$, and $\yxvec = 0$ if $\xvec$ is a normal sample otherwise; and $g_p(\mathbf{M}_{\xvec};\Theta_p)$ is exactly the same as $g_s$ in Eq. \eqref{eqn:seenabnormality}, but $g_p$ is trained in a separate head with different anomaly data and parameters from $g_s$
to learn the pseudo abnormality. As discussed in Secs. \ref{sec:implementation} and \ref{sec:sensitivity}, the outlier exposure method \cite{hendrycks2018deep} is used in anomaly detection on medical datasets. In such cases, the pseudo anomalies $\tilde{\mathbf{x}}$ are samples randomly drawn from external data instead of creating from Eq. \eqref{eqn:dataaugmentation}.

\textbf{Abnormality Learning with Latent Residual Anomalies}.
Some anomalies, such as previously unknown anomalies that share no common abnormal features with the seen anomalies and have only small difference to the normal samples, are difficult to detect by using only the features of the anomalies themselves, but they can be easily detected in a high-order composite feature space provided that the composite features are more discriminative. As anomalies are characterized by their difference from normal data, we utilize the difference between the features of the anomalies and normal feature representations to learn such discriminative composite features.
More specifically, we propose the latent residual abnormality learning that learns anomaly scores of samples based on their feature residuals comparing to the features of some reference images (normal images) in a learned feature space. As shown in Fig. \ref{fig:c}, to obtain the latent feature residuals, we first use a small set of images randomly drawn from the normal data as the reference data, and compute the mean of their feature maps to obtain the reference normal feature map:
\begin{equation}
    \mathbf{M}_r = \frac{1}{\mathit{N}_{r}}\sum_{i=1}^{\mathit{N}_{r}} f(\xvec_{{r}_i};\Theta_f),
\end{equation}
where $\xvec_{{r}_i}$ is a reference normal image, and $\mathit{N}_{r}$ is a hyperparameter that represents the size of the reference set. 
For a given training image $\xvec$, we perform element-wise subtraction between its feature map $\mathbf{M}_{\xvec}$ and the reference normal feature map $\mathbf{M}_r$ that is fixed for all training and testing samples, resulting in a residual feature map $\mathbf{M}_{r \ominus \xvec}$ for $\xvec$:
\begin{equation}
    \mathbf{M}_{r \ominus \xvec} = \mathbf{M}_r \ominus  \mathbf{M}_{\xvec},
\end{equation}
where $\ominus$ denotes element-wise subtraction. We then perform an anomaly classification upon these residual features:
\begin{equation}
    \ell_{r}(\mathbf{x}, \yxvec) = \ell\big(g_r(\mathbf{M}_{r \ominus \xvec};\Theta_r), \yxvec\big),
\end{equation}
where $\yxvec = 1$ if $\xvec$ is a seen/pseudo anomaly, and $\yxvec = 0$ if $\xvec$ is a normal sample otherwise. Again, $g_r$ uses exactly the same method to obtain the anomaly score as $g_s$ in Eq. \ref{eqn:seenabnormality}, but it is trained in a separate head with the parameters $\Theta_r$ using different training inputs, \ie, residual feature map $\mathbf{M}_{r \ominus \xvec}$.

Since the $g_s$, $g_p$ and $g_r$ heads focus on learning the abnormality representations, the jointly learned feature map in $f$ does not well model the normal features. To address this issue, we add a separate normality learning head as follows:
\begin{equation}
    \ell_{n}(\mathbf{x}, \yxvec) = \ell\Big(g_n\big(\frac{1}{h^\prime\times w^\prime}\sum_{i=1}^{h^\prime \times w^\prime} \mathbf{d}_i; \Theta_{n}\big), \yxvec\Big),
\end{equation}
where $g_n: \mathcal{D}\rightarrow\mathbb{R}$ is a fully-connected binary anomaly classifier that discriminates normal samples from all seen and pseudo anomalies. Unlike abnormal features that are often fine-grained local features, normal features are holistic global features. Hence, 
$g_n$ does not use the top-$K$ MIL-based anomaly scoring as in other heads and learns holistic normal scores instead.

\textbf{Training and Inference}. During training, the feature mapping network $f$ is shared and jointly trained by all the four heads $g_s$, $g_p$, $g_r$ and $g_n$. These four heads are independent from each other, and so their parameters are not shared and independently optimized. A loss function called deviation loss \cite{pang2019deep,pang2021explainable} is used to implement the loss function $\ell$ in all our heads by default, as it enables generally more stable and effective performance than other loss functions such as cross entropy loss or focal loss (see \texttt{Appendix C.2}). During inference, given a test image, we sum of all the scores from the abnormality learning heads ($g_s$, $g_p$ and $g_r$) and minus the score from the normality head $g_n$ to obtain its anomaly score.
\section{Experiments}

\begin{table*}[bt]
  \centering
  \caption{AUC results (mean±std) on nine real-world AD datasets
  under the general setting. The first 15 datasets are data subsets of MVTec AD whose results are the averaged results over these subsets. The supervised methods are trained using one or ten random anomaly examples, with the best results in \textcolor{red}{\textbf{red}} and the second-best in \textcolor{blue}{\textbf{blue}}. KDAD is treated as a baseline. $|\mathcal{C}|$ is the number of anomaly classes.
}
  \vspace{-0.2cm}
  \scalebox{0.66}{
    \begin{tabular}{l@{}|c ||c||ccccc||ccccc}
    \hline
    \multirow{2}{*}{  \textbf{Dataset} } & \multirow{2}{*}{$|\mathcal{C}|$}  &    \textbf{Baseline} & \multicolumn{5}{c||}{\textbf{One Training Anomaly Example}} & \multicolumn{5}{c}{\textbf{Ten Training Anomaly Examples}}\\
    \cline{3-13}
    & & \multicolumn{1}{c||}{\textbf{KDAD}} & \multicolumn{1}{c}{\textbf{DevNet}} & \multicolumn{1}{c}{\textbf{FLOS}} & \multicolumn{1}{c}{\textbf{SAOE}} & \multicolumn{1}{c}{\textbf{MLEP}} & \multicolumn{1}{c||}{\textbf{DRA (Ours)}}& \multicolumn{1}{c}{\textbf{DevNet}} & \multicolumn{1}{c}{\textbf{FLOS}} & \multicolumn{1}{c}{\textbf{SAOE}} & \multicolumn{1}{c}{\textbf{MLEP}} & \multicolumn{1}{c}{\textbf{DRA (Ours)}} \\\hline
    Carpet & 5 & 0.774\footnotesize{±0.005}& 0.746\footnotesize{±0.076}& 0.755\footnotesize{±0.026}& \textcolor{blue}{\textbf{0.766}}\footnotesize{±0.098}& 0.701\footnotesize{±0.091}& \textcolor{red}{\textbf{0.859}}\footnotesize{±0.023}& \textcolor{blue}{\textbf{0.867}}\footnotesize{±0.040}& 0.780\footnotesize{±0.009}& 0.755\footnotesize{±0.136}& 0.781\footnotesize{±0.049}& \textcolor{red}{\textbf{0.940}}\footnotesize{±0.027} \\
    
    Grid & 5 & 0.749\footnotesize{±0.017}& 0.891\footnotesize{±0.040}& 0.871\footnotesize{±0.076}& \textcolor{blue}{\textbf{0.921}}\footnotesize{±0.032}& 0.839\footnotesize{±0.028}& \textcolor{red}{\textbf{0.972}}\footnotesize{±0.011}& 0.967\footnotesize{±0.021}& 0.966\footnotesize{±0.005}& 0.952\footnotesize{±0.011}& \textcolor{blue}{\textbf{0.980}}\footnotesize{±0.009}& \textcolor{red}{\textbf{0.987}}\footnotesize{±0.009}  \\
    
    Leather & 5& 0.948\footnotesize{±0.005}& 0.873\footnotesize{±0.026}& 0.791\footnotesize{±0.057}& \textcolor{red}{\textbf{0.996}}\footnotesize{±0.007}& 0.781\footnotesize{±0.020}& \textcolor{blue}{\textbf{0.989}}\footnotesize{±0.005}& \textcolor{blue}{\textbf{0.999}}\footnotesize{±0.001}& 0.993\footnotesize{±0.004}& \textcolor{red}{\textbf{1.000}}\footnotesize{±0.000}& 0.813\footnotesize{±0.158}& \textcolor{red}{\textbf{1.000}}\footnotesize{±0.000} \\
    
    Tile & 5& 0.911\footnotesize{±0.010}& 0.752\footnotesize{±0.038}& 0.787\footnotesize{±0.038}& \textcolor{blue}{\textbf{0.935}}\footnotesize{±0.034}& 0.927\footnotesize{±0.036}& \textcolor{red}{\textbf{0.965}}\footnotesize{±0.015}& 0.987\footnotesize{±0.005}& 0.952\footnotesize{±0.010}& 0.944\footnotesize{±0.013}& \textcolor{blue}{\textbf{0.988}}\footnotesize{±0.009}& \textcolor{red}{\textbf{0.994}}\footnotesize{±0.006} \\
    
    Wood & 5& 0.940\footnotesize{±0.004}& 0.900\footnotesize{±0.068}& 0.927\footnotesize{±0.065}& \textcolor{blue}{\textbf{0.948}}\footnotesize{±0.009}& 0.660\footnotesize{±0.142}& \textcolor{red}{\textbf{0.985}}\footnotesize{±0.011}& \textcolor{blue}{\textbf{0.999}}\footnotesize{±0.001}& \textcolor{red}{\textbf{1.000}}\footnotesize{±0.000}& 0.976\footnotesize{±0.031}& \textcolor{blue}{\textbf{0.999}}\footnotesize{±0.002}& 0.998\footnotesize{±0.001} \\

    Bottle & 3& 0.992\footnotesize{±0.002}& 0.976\footnotesize{±0.006}& 0.975\footnotesize{±0.023}& \textcolor{blue}{\textbf{0.989}}\footnotesize{±0.019}& 0.927\footnotesize{±0.090}& \textcolor{red}{\textbf{1.000}}\footnotesize{±0.000}& 0.993\footnotesize{±0.008}& 0.995\footnotesize{±0.002}& \textcolor{blue}{\textbf{0.998}}\footnotesize{±0.003}& 0.981\footnotesize{±0.004}& \textcolor{red}{\textbf{1.000}}\footnotesize{±0.000} \\
    
    Capsule & 5& 0.775\footnotesize{±0.019}& 0.564\footnotesize{±0.032}& \textcolor{red}{\textbf{0.666}}\footnotesize{±0.020}& 0.611\footnotesize{±0.109}& 0.558\footnotesize{±0.075}& \textcolor{blue}{\textbf{0.631}}\footnotesize{±0.056}& 0.865\footnotesize{±0.057}& \textcolor{blue}{\textbf{0.902}}\footnotesize{±0.017}& 0.850\footnotesize{±0.054}& 0.818\footnotesize{±0.063}& \textcolor{red}{\textbf{0.935}}\footnotesize{±0.022} \\
    
    Pill & 7& 0.824\footnotesize{±0.006}& \textcolor{blue}{\textbf{0.769}}\footnotesize{±0.017}& 0.745\footnotesize{±0.064}& 0.652\footnotesize{±0.078}& 0.656\footnotesize{±0.061}& \textcolor{red}{\textbf{0.832}}\footnotesize{±0.034}& 0.866\footnotesize{±0.038}& \textcolor{red}{\textbf{0.929}}\footnotesize{±0.012}& 0.872\footnotesize{±0.049}& 0.845\footnotesize{±0.048}& \textcolor{blue}{\textbf{0.904}}\footnotesize{±0.024} \\
    
    Transistor & 4& 0.805\footnotesize{±0.013}& \textcolor{red}{\textbf{0.722}}\footnotesize{±0.032}& \textcolor{blue}{\textbf{0.709}}\footnotesize{±0.041}& 0.680\footnotesize{±0.182}& 0.695\footnotesize{±0.124}& 0.668\footnotesize{±0.068}& \textcolor{blue}{\textbf{0.924}}\footnotesize{±0.027}& 0.862\footnotesize{±0.037}& 0.860\footnotesize{±0.053}& \textcolor{red}{\textbf{0.927}}\footnotesize{±0.043}& 0.915\footnotesize{±0.025} \\
    
    Zipper & 7& 0.927\footnotesize{±0.018}& 0.922\footnotesize{±0.018}& 0.885\footnotesize{±0.033}& \textcolor{blue}{\textbf{0.970}}\footnotesize{±0.033}& 0.856\footnotesize{±0.086}& \textcolor{red}{\textbf{0.984}}\footnotesize{±0.016}& 0.990\footnotesize{±0.009}& 0.990\footnotesize{±0.008}& \textcolor{blue}{\textbf{0.995}}\footnotesize{±0.004}& 0.965\footnotesize{±0.002}& \textcolor{red}{\textbf{1.000}}\footnotesize{±0.000} \\
    
    Cable &8& 0.880\footnotesize{±0.002}& 0.783\footnotesize{±0.058}& 0.790\footnotesize{±0.039}& \textcolor{blue}{\textbf{0.819}}\footnotesize{±0.060}& 0.688\footnotesize{±0.017}& \textcolor{red}{\textbf{0.876}}\footnotesize{±0.012}& \textcolor{blue}{\textbf{0.892}}\footnotesize{±0.020}& 0.890\footnotesize{±0.063}& 0.862\footnotesize{±0.022}& 0.857\footnotesize{±0.062}& \textcolor{red}{\textbf{0.909}}\footnotesize{±0.011} \\
    
    Hazelnut &4& 0.984\footnotesize{±0.001}& \textcolor{red}{\textbf{0.979}}\footnotesize{±0.010}& 0.976\footnotesize{±0.021}& 0.961\footnotesize{±0.042}& 0.704\footnotesize{±0.090}& \textcolor{blue}{\textbf{0.977}}\footnotesize{±0.030}& \textcolor{red}{\textbf{1.000}}\footnotesize{±0.000}& \textcolor{red}{\textbf{1.000}}\footnotesize{±0.000}& \textcolor{red}{\textbf{1.000}}\footnotesize{±0.000}& \textcolor{red}{\textbf{1.000}}\footnotesize{±0.000}& \textcolor{red}{\textbf{1.000}}\footnotesize{±0.000} \\
    
    Metal\_nut & 4 & 0.743\footnotesize{±0.013}& 0.876\footnotesize{±0.007}& \textcolor{blue}{\textbf{0.930}}\footnotesize{±0.022}& 0.922\footnotesize{±0.033}& 0.878\footnotesize{±0.038}& \textcolor{red}{\textbf{0.948}}\footnotesize{±0.046}& \textcolor{blue}{\textbf{0.991}}\footnotesize{±0.006}& 0.984\footnotesize{±0.004}& 0.976\footnotesize{±0.013}& 0.974\footnotesize{±0.009}& \textcolor{red}{\textbf{0.997}}\footnotesize{±0.002} \\
    
    Screw & 5& 0.805\footnotesize{±0.021}& 0.399\footnotesize{±0.187}& 0.337\footnotesize{±0.091}& 0.653\footnotesize{±0.074}& \textcolor{blue}{\textbf{0.675}}\footnotesize{±0.294}& \textcolor{red}{\textbf{0.903}}\footnotesize{±0.064}& 0.970\footnotesize{±0.015}& 0.940\footnotesize{±0.017}& \textcolor{blue}{\textbf{0.975}}\footnotesize{±0.023}& 0.899\footnotesize{±0.039}& \textcolor{red}{\textbf{0.977}}\footnotesize{±0.009} \\
    
    Toothbrush & 1 & 0.863\footnotesize{±0.029}& \textcolor{red}{\textbf{0.753}}\footnotesize{±0.027}& \textcolor{blue}{\textbf{0.731}}\footnotesize{±0.028}& 0.686\footnotesize{±0.110}& 0.617\footnotesize{±0.058}& 0.650\footnotesize{±0.029}& 0.860\footnotesize{±0.066}& \textcolor{red}{\textbf{0.900}}\footnotesize{±0.008}& \textcolor{blue}{\textbf{0.865}}\footnotesize{±0.062}& 0.783\footnotesize{±0.048}& 0.826\footnotesize{±0.021} \\
    
    \hline
    \textbf{MVTec AD} & - & 0.861\footnotesize{±0.009}& 0.794\footnotesize{±0.014}& 0.792\footnotesize{±0.014}& \textcolor{blue}{\textbf{0.834}}\footnotesize{±0.007}& 0.744\footnotesize{±0.019}& \textcolor{red}{\textbf{0.883}}\footnotesize{±0.008}& \textcolor{blue}{\textbf{0.945}}\footnotesize{±0.004}& 0.939\footnotesize{±0.007}& 0.926\footnotesize{±0.010}& 0.907\footnotesize{±0.005}& \textcolor{red}{\textbf{0.959}}\footnotesize{±0.003} \\
    
    \textbf{AITEX } &12& 0.576\footnotesize{±0.002}& 0.598\footnotesize{±0.070}& 0.538\footnotesize{±0.073}& \textcolor{blue}{\textbf{0.675}}\footnotesize{±0.094}& 0.564\footnotesize{±0.055}& \textcolor{red}{\textbf{0.692}}\footnotesize{±0.124}& \textcolor{blue}{\textbf{0.887}}\footnotesize{±0.013}& 0.841\footnotesize{±0.049}& 0.874\footnotesize{±0.024}& 0.867\footnotesize{±0.037}& \textcolor{red}{\textbf{0.893}}\footnotesize{±0.017} \\
    
    \textbf{SDD}& 1 & 0.888\footnotesize{±0.005}& \textcolor{red}{\textbf{0.881}}\footnotesize{±0.009}& 0.840\footnotesize{±0.043}& 0.781\footnotesize{±0.009}& 0.811\footnotesize{±0.045}& \textcolor{blue}{\textbf{0.859}}\footnotesize{±0.014}& \textcolor{blue}{\textbf{0.988}}\footnotesize{±0.006}& 0.967\footnotesize{±0.018}& 0.955\footnotesize{±0.020}& 0.983\footnotesize{±0.013}& \textcolor{red}{\textbf{0.991}}\footnotesize{±0.005} \\
    
    \textbf{ELPV} & 2 & 0.744\footnotesize{±0.001}& 0.514\footnotesize{±0.076}& 0.457\footnotesize{±0.056}& \textcolor{blue}{\textbf{0.635}}\footnotesize{±0.092}& 0.578\footnotesize{±0.062}& \textcolor{red}{\textbf{0.675}}\footnotesize{±0.024}& \textcolor{red}{\textbf{0.846}}\footnotesize{±0.022}& 0.818\footnotesize{±0.032}& 0.793\footnotesize{±0.047}& 0.794\footnotesize{±0.047}& \textcolor{blue}{\textbf{0.845}}\footnotesize{±0.013} \\
    
   \textbf{Optical}& 1 & 0.579\footnotesize{±0.002}& 0.523\footnotesize{±0.003}& 0.518\footnotesize{±0.003}& \textcolor{blue}{\textbf{0.815}}\footnotesize{±0.014}& 0.516\footnotesize{±0.009}& \textcolor{red}{\textbf{0.888}}\footnotesize{±0.012}& 0.782\footnotesize{±0.065}& 0.720\footnotesize{±0.055}& \textcolor{blue}{\textbf{0.941}}\footnotesize{±0.013}& 0.740\footnotesize{±0.039}& \textcolor{red}{\textbf{0.965}}\footnotesize{±0.006}\\
   
   \textbf{Mastcam}& 11& 0.642\footnotesize{±0.007}& 0.595\footnotesize{±0.016}& 0.542\footnotesize{±0.017}& \textcolor{blue}{\textbf{0.662}}\footnotesize{±0.018}& 0.625\footnotesize{±0.045}& \textcolor{red}{\textbf{0.692}}\footnotesize{±0.058}& 0.790\footnotesize{±0.021}& 0.703\footnotesize{±0.029}& \textcolor{blue}{\textbf{0.810}}\footnotesize{±0.029}& 0.798\footnotesize{±0.026}& \textcolor{red}{\textbf{0.848}}\footnotesize{±0.008}\\
   
    \textbf{BrainMRI}& 1 & 0.733\footnotesize{±0.016}& \textcolor{blue}{\textbf{0.694}}\footnotesize{±0.004}& 0.693\footnotesize{±0.036}& 0.531\footnotesize{±0.060}& 0.632\footnotesize{±0.017}& \textcolor{red}{\textbf{0.744}}\footnotesize{±0.004}& 0.958\footnotesize{±0.012}& 0.955\footnotesize{±0.011}& 0.900\footnotesize{±0.041}& \textcolor{blue}{\textbf{0.959}}\footnotesize{±0.011}& \textcolor{red}{\textbf{0.970}}\footnotesize{±0.003}\\
    
   \textbf{HeadCT}& 1 & 0.793\footnotesize{±0.017}& 0.742\footnotesize{±0.076}& 0.698\footnotesize{±0.092}& 0.597\footnotesize{±0.022}& \textcolor{blue}{\textbf{0.758}}\footnotesize{±0.038}& \textcolor{red}{\textbf{0.796}}\footnotesize{±0.105}& \textcolor{red}{\textbf{0.982}}\footnotesize{±0.009}& 0.971\footnotesize{±0.004}& 0.935\footnotesize{±0.021}& \textcolor{blue}{\textbf{0.972}}\footnotesize{±0.014}& \textcolor{blue}{\textbf{0.972}}\footnotesize{±0.002}\\
   
   \textbf{Hyper-Kvasir$\ $}& 4 & 0.401\footnotesize{±0.002}& 0.653\footnotesize{±0.037}& \textcolor{blue}{\textbf{0.668}}\footnotesize{±0.004}& 0.498\footnotesize{±0.100}& 0.445\footnotesize{±0.040}& \textcolor{red}{\textbf{0.690}}\footnotesize{±0.017}& \textcolor{blue}{\textbf{0.829}}\footnotesize{±0.018}& 0.773\footnotesize{±0.029}& 0.666\footnotesize{±0.050}& 0.600\footnotesize{±0.069}& \textcolor{red}{\textbf{0.834}}\footnotesize{±0.004}\\
   \hline
    \end{tabular}
    }
  \label{tab:randomanomalies}%
  \vspace{-0.3cm}
\end{table*}%

\textbf{Datasets}
Many studies 
evaluate their models on synthetic anomaly detection datasets converted from popular image classification benchmarks, such as MNIST \cite{lecun1998gradient}, Fashion-MNIST \cite{xiao2017_online}, CIFAR-10 \cite{krizhevsky2009learning}, using one-vs-all or one-vs-one protocols. This conversion results in clearly disparate anomalies from normal samples. However, anomalies and normal samples in real-world applications, such as industrial defect inspection and lesion detection in medical images, typically have only subtle/small difference. Motivated by this, following \cite{yi2020patch, li2021cutpaste, pang2021explainable
}, we focus on datasets with natural anomalies rather than one-vs-all/one-vs-one based synthetic anomalies. Particularly, nine diverse datasets with real anomalies are used in our experiments, including five industrial defect inspection datasets: \textbf{MVTec AD} \cite{Bergmann_2019_CVPR}, \textbf{AITEX} \cite{silvestre2019public}, \textbf{SDD} \cite{Tabernik2019JIM}, \textbf{ELPV} \cite{deitsch2019elpv} and \textbf{Optical} \cite{wieler2007weakly}, in which we aim to inspect defective image samples; one planetary exploration dataset: \textbf{Mastcam} \cite{kerner2020comparison} in which we aim to identify geologically-interesting/novel images taken by Mars exploration rovers; and three medical image datasets for detecting lesions on different organs: \textbf{BrainMRI} \cite{salehi2021multiresolution}, \textbf{HeadCT} \cite{salehi2021multiresolution} and \textbf{Hyper-Kvasir} \cite{borgli2020hyperkvasir}. These datasets are popular benchmarks in the respective research domains and recently emerging as important benchmarks for anomaly detection \cite{Bergmann_2020_CVPR,yi2020patch,salehi2021multiresolution,pang2021explainable, Hou_2021_ICCV} (see \texttt{Appendix A} for detailed introduction of these datasets).

\subsection{Implementation Details}\label{sec:implementation}

DRA uses ResNet-18 as the feature learning backbone. 
All its heads are jointly trained using 30 epochs, with 20 iterations per epoch and a batch size of 48. Adam is used for the parameter optimization using an initial learning rate $10^{-3}$ with a weight decay of $10^{-2}$. The top-$K$ MIL in DRA is the same as that in DevNet \cite{pang2021explainable}, \ie, $K$ in the top-$K$ MIL is set to 10\% of the number of all scores per score map. $\mathit{N}_{r}=5$ is used by default in the residual anomaly learning (see Sec. \ref{sec:sensitivity}).
The pseudo abnormality learning uses CutMix \cite{yun2019cutmix} to create pseudo anomaly samples on all datasets except the three medical datasets, on which DRA uses external data from another medical dataset LAG \cite{Li_2019_CVPR} as the pseudo anomaly source (see Sec. \ref{sec:sensitivity}). 

Our model DRA is compared to five recent and closely related state-of-the-art (SotA) methods, including MLEP \cite{liu2019margin}, deviation network (DevNet) \cite{pang2021explainable,pang2019deep}, SAOE (combining data augmentation-based \underline{S}ynthetic \underline{A}nomalies \cite{li2021cutpaste, liznerski2021explainable,tack2020csi} with \underline{O}utlier \underline{E}xposure \cite{hendrycks2018deep, reiss2021panda}), unsupervised anomaly detector KDAD \cite{salehi2021multiresolution}, and focal loss-driven classifier (FLOS) \cite{lin2017focalloss} (See \texttt{Appendix C.1} for comparison with two other methods \cite{ruff2019deep,wang2018revisiting}). 
MLEP and DevNet address the same open-set AD problem as ours. KDAD is a recent unsupervised AD method that works on normal training data only. It is commonly assumed that unsupervised detectors are more preferable than the supervised ones in detecting unseen anomalies, as the latter may bias towards the seen anomalies. Motivated by this, KDAD is used as a baseline. The implementation of DevNet and KDAD is taken from their authors. MLEP is adapted to the image task with the same setting as DRA. SAOE utilizes pseudo anomalies from both data augmentation-based and outlier exposure-based methods, outperforming the individuals that use one of these anomaly creation methods.
FLOS is an imbalanced classifier trained with focal loss. 
For a fair comparison, all competing methods use the same network backbone (\ie, ResNet-18) as DRA except KDAD that requires its own special network architecture to perform training and inference. Further implementation details of DRA and its competing methods are provided in \texttt{Appendix B}.

\subsection{Experiment Protocols}
We use the following two experiment protocols:

\textbf{General setting} simulates a general scenario of open-set AD, where the given anomaly examples are a few samples randomly drawn from all possible anomaly classes in the test set per dataset. These sampled anomalies are then removed from the test data. This is to replicate real-world applications where we cannot determine which anomaly classes are known and how many anomaly classes the given anomaly examples span. Thus, the datasets can contain both seen and unseen anomaly classes, or only the seen anomaly classes, depending on the underlying complexity of the applications (\eg, the number of all possible anomaly classes).

\textbf{Hard setting} is designed to exclusively evaluate the performance of the models in detecting unseen anomaly classes, which is the very key challenge in open-set AD. To this end,
the anomaly example sampling is limited to be drawn from one single anomaly class only, and all anomaly samples in this anomaly class are removed from the test set to ensure that the test set contains only unseen anomaly classes. 
Note that this setting is only applicable to datasets with no less than two anomaly classes.

As labeled anomalies are difficult to obtain due to their rareness and unknowingness, in both settings we use only very limited labeled anomalies, \ie, with the number of the given anomaly examples respectively fixed to one and ten. 
The popular performance metric, Area Under ROC Curve (AUC), is used. Each model yields an anomaly ranking, and its AUC is calculated based on the ranking. All reported AUCs are averaged results over three independent runs.

\subsection{Results under the General Setting}
Tab. \ref{tab:randomanomalies} shows the comparison results 
under the general setting protocol. Below we discuss the results in details.

\textbf{Application Domain Perspective}. Despite the datasets from diverse application domains, including industrial defect inspection, rover-based planetary exploration and medical image analysis, our model achieves the best AUC performance on across nearly all of the datasets, \ie, eight (seven) out of nine datasets in the one-shot (ten-shot) setting, with the second-best results on the other datasets. On challenging datasets, such as MVTec AD, AITEX, Mastcam and Hyper-Kvasir, where a larger number of possible anomaly classes is presented, our model obtains consistently better AUC results, increasing by up to 5\% AUC.

\textbf{Sample Efficiency}. The reduction of training anomaly examples generally decreases the performance of all the supervised models. Compared to the competing detectors, our model shows better sample efficiency in that i) with reduced anomaly examples, our model has a much smaller decrease of AUC, \ie, an average of 15.1\% AUC decrease across the nine datasets, which is much better than DevNet (22.3\%), FLOS (21.6\%), SAOE (19.7\%), and MLEP (21.6\%), and ii) our model trained with one anomaly example can largely outperform the strong competing methods trained with ten anomaly examples, such as DevNet, FLOS and MLEP on Optical, and SAOE and MLEP on Hyper-Kvasir.

\textbf{Comparison to Unsupervised Baseline}. Compared to the unsupervised model KDAD, our model and other supervised models demonstrate consistently better performance when using ten training anomaly examples (\ie, less open-set scenarios). In more open-set scenarios where only one anomaly example is used, our method is the only model that is still clearly better than KDAD on most datasets, even on challenging datasets which have many anomaly classes, such as MVTec AD, AITEX, and Mastcam. 

\subsection{Results under the Hard Setting}

\begin{table*}[bt]
  \centering
  \caption{AUC results under the hard setting, where models are trained with one known anomaly class and tested to detect the rest of all other anomaly classes. Each data subset is named by the known anomaly class.}
  \vspace{-0.2cm}
  \scalebox{0.63}{
    \begin{tabular}{l@{}l@{}|c||ccccc||ccccc}
    \hline
    \multirow{2}{*}{\textbf{Dataset}}      & \multirow{2}{*}{\textbf{Data Subset}} & \textbf{Baseline}  & \multicolumn{5}{c||}{\textbf{One Training Anomaly Example}} & \multicolumn{5}{c}{\textbf{Ten Training Anomaly Examples}}\\
    \cline{3-13} 
    && \multicolumn{1}{c||}{\textbf{KDAD}} & \multicolumn{1}{c}{\textbf{DevNet}} & \multicolumn{1}{c}{\textbf{FLOS}} & \multicolumn{1}{c}{\textbf{SAOE}} & \multicolumn{1}{c}{\textbf{MLEP}} & \multicolumn{1}{c||}{\textbf{DRA (Ours)}}& \multicolumn{1}{c}{\textbf{DevNet}} & \multicolumn{1}{c}{\textbf{FLOS}} & \multicolumn{1}{c}{\textbf{SAOE}} & \multicolumn{1}{c}{\textbf{MLEP}} & \multicolumn{1}{c}{\textbf{DRA (Ours)}} \\
    \hline
    \multirow{6}{*}{\textbf{Carpet}} & Color & 0.787\footnotesize{±0.005}& 0.716\footnotesize{±0.085}& 0.467\footnotesize{±0.278}& \textcolor{blue}{\textbf{0.763}}\footnotesize{±0.100}& 0.547\footnotesize{±0.056}& \textcolor{red}{\textbf{0.879}}\footnotesize{±0.021}& \textcolor{blue}{\textbf{0.767}}\footnotesize{±0.015}& 0.760\footnotesize{±0.005}& 0.467\footnotesize{±0.067}& 0.698\footnotesize{±0.025}& \textcolor{red}{\textbf{0.886}}\footnotesize{±0.042} \\
          & Cut & 0.766\footnotesize{±0.005}& 0.666\footnotesize{±0.035}& \textcolor{blue}{\textbf{0.685}}\footnotesize{±0.007}& 0.664\footnotesize{±0.165}& 0.658\footnotesize{±0.056}& \textcolor{red}{\textbf{0.902}}\footnotesize{±0.033}& \textcolor{blue}{\textbf{0.819}}\footnotesize{±0.037}& 0.688\footnotesize{±0.059}& 0.793\footnotesize{±0.175}& 0.653\footnotesize{±0.120}& \textcolor{red}{\textbf{0.922}}\footnotesize{±0.038}  \\
          & Hole  & 0.757\footnotesize{±0.003}& 0.721\footnotesize{±0.067}& 0.594\footnotesize{±0.142}& \textcolor{blue}{\textbf{0.772}}\footnotesize{±0.071}& 0.653\footnotesize{±0.065}& \textcolor{red}{\textbf{0.901}}\footnotesize{±0.033}& 0.814\footnotesize{±0.038}& 0.733\footnotesize{±0.014}& \textcolor{blue}{\textbf{0.831}}\footnotesize{±0.125}& 0.674\footnotesize{±0.076}& \textcolor{red}{\textbf{0.947}}\footnotesize{±0.016} \\
          & Metal & 0.836\footnotesize{±0.003}& \textcolor{blue}{\textbf{0.819}}\footnotesize{±0.032}& 0.701\footnotesize{±0.028}& 0.780\footnotesize{±0.172}& 0.706\footnotesize{±0.047}& \textcolor{red}{\textbf{0.871}}\footnotesize{±0.037}& 0.863\footnotesize{±0.022}& 0.678\footnotesize{±0.083}& \textcolor{blue}{\textbf{0.883}}\footnotesize{±0.043}& 0.764\footnotesize{±0.061}& \textcolor{red}{\textbf{0.933}}\footnotesize{±0.022} \\
          & Thread & 0.750\footnotesize{±0.005}& 0.912\footnotesize{±0.044}& \textcolor{blue}{\textbf{0.941}}\footnotesize{±0.005}& 0.787\footnotesize{±0.204}& 0.831\footnotesize{±0.117}& \textcolor{red}{\textbf{0.950}}\footnotesize{±0.029}& \textcolor{blue}{\textbf{0.972}}\footnotesize{±0.009}& 0.946\footnotesize{±0.005}& 0.834\footnotesize{±0.297}& 0.967\footnotesize{±0.006}& \textcolor{red}{\textbf{0.989}}\footnotesize{±0.004} \\
          \cline{2-13}
          & \textbf{Mean} & 0.779\footnotesize{±0.002}& \textcolor{blue}{\textbf{0.767}}\footnotesize{±0.018}& 0.678\footnotesize{±0.040}& 0.753\footnotesize{±0.055}& 0.679\footnotesize{±0.029}& \textcolor{red}{\textbf{0.901}}\footnotesize{±0.006}& \textcolor{blue}{\textbf{0.847}}\footnotesize{±0.017}& 0.761\footnotesize{±0.012}& 0.762\footnotesize{±0.073}& 0.751\footnotesize{±0.023}& \textcolor{red}{\textbf{0.935}}\footnotesize{±0.013} \\
    \hline
    \multirow{5}[0]{*}{\textbf{Metal\_nut}} & Bent & 0.798\footnotesize{±0.015}& 0.797\footnotesize{±0.048}& 0.851\footnotesize{±0.046}& \textcolor{blue}{\textbf{0.864}}\footnotesize{±0.032}& 0.743\footnotesize{±0.013}& \textcolor{red}{\textbf{0.952}}\footnotesize{±0.020}& 0.904\footnotesize{±0.022}& 0.827\footnotesize{±0.075}& 0.901\footnotesize{±0.023}& \textcolor{blue}{\textbf{0.956}}\footnotesize{±0.013}& \textcolor{red}{\textbf{0.990}}\footnotesize{±0.003} \\
          & Color & 0.754\footnotesize{±0.014}& \textcolor{blue}{\textbf{0.909}}\footnotesize{±0.023}& 0.821\footnotesize{±0.059}& 0.857\footnotesize{±0.037}& 0.835\footnotesize{±0.075}& \textcolor{red}{\textbf{0.946}}\footnotesize{±0.023}& \textcolor{red}{\textbf{0.978}}\footnotesize{±0.016}& \textcolor{red}{\textbf{0.978}}\footnotesize{±0.008}& 0.879\footnotesize{±0.018}& 0.945\footnotesize{±0.039}& \textcolor{blue}{\textbf{0.967}}\footnotesize{±0.011} \\
          & Flip & 0.646\footnotesize{±0.019}& 0.764\footnotesize{±0.014}& 0.799\footnotesize{±0.058}& 0.751\footnotesize{±0.090}& \textcolor{blue}{\textbf{0.813}}\footnotesize{±0.031}& \textcolor{red}{\textbf{0.921}}\footnotesize{±0.029}& \textcolor{red}{\textbf{0.987}}\footnotesize{±0.004}& \textcolor{blue}{\textbf{0.942}}\footnotesize{±0.009}& 0.795\footnotesize{±0.062}& 0.805\footnotesize{±0.057}& 0.913\footnotesize{±0.021} \\
          & Scratch & 0.737\footnotesize{±0.010}& \textcolor{red}{\textbf{0.952}}\footnotesize{±0.052}& 0.947\footnotesize{±0.027}& 0.792\footnotesize{±0.075}& 0.907\footnotesize{±0.085}& \textcolor{blue}{\textbf{0.909}}\footnotesize{±0.023}& \textcolor{red}{\textbf{0.991}}\footnotesize{±0.017}& \textcolor{blue}{\textbf{0.943}}\footnotesize{±0.002}& 0.845\footnotesize{±0.041}& 0.805\footnotesize{±0.153}& 0.911\footnotesize{±0.034} \\
          \cline{2-13}
          & \textbf{Mean} & 0.734\footnotesize{±0.005}& \textcolor{blue}{\textbf{0.855}}\footnotesize{±0.016}& \textcolor{blue}{\textbf{0.855}}\footnotesize{±0.024}& 0.816\footnotesize{±0.029}& 0.825\footnotesize{±0.023}& \textcolor{red}{\textbf{0.932}}\footnotesize{±0.017}& \textcolor{red}{\textbf{0.965}}\footnotesize{±0.011}& 0.922\footnotesize{±0.014}& 0.855\footnotesize{±0.016}& 0.878\footnotesize{±0.058}& \textcolor{blue}{\textbf{0.945}}\footnotesize{±0.017}  \\
    \hline
    \multirow{7}[0]{*}{\textbf{AITEX}} & Broken\_end & 0.552\footnotesize{±0.006}& \textcolor{blue}{\textbf{0.712}}\footnotesize{±0.069}& 0.645\footnotesize{±0.030}& \textcolor{red}{\textbf{0.778}}\footnotesize{±0.068}& 0.441\footnotesize{±0.111}& 0.708\footnotesize{±0.094}& 0.658\footnotesize{±0.111}& 0.585\footnotesize{±0.037}& \textcolor{blue}{\textbf{0.712}}\footnotesize{±0.068}& \textcolor{red}{\textbf{0.732}}\footnotesize{±0.065}& 0.693\footnotesize{±0.099} \\
          & Broken\_pick & 0.705\footnotesize{±0.003}& 0.552\footnotesize{±0.003}& 0.598\footnotesize{±0.023}& \textcolor{blue}{\textbf{0.644}}\footnotesize{±0.039}& 0.476\footnotesize{±0.070}& \textcolor{red}{\textbf{0.731}}\footnotesize{±0.072}& 0.585\footnotesize{±0.028}& 0.548\footnotesize{±0.054}& \textcolor{blue}{\textbf{0.629}}\footnotesize{±0.012}& 0.555\footnotesize{±0.027}& \textcolor{red}{\textbf{0.760}}\footnotesize{±0.037} \\
          & Cut\_selvage & 0.567\footnotesize{±0.006}& 0.689\footnotesize{±0.016}& \textcolor{blue}{\textbf{0.694}}\footnotesize{±0.036}& 0.681\footnotesize{±0.077}& 0.434\footnotesize{±0.149}& \textcolor{red}{\textbf{0.739}}\footnotesize{±0.101}& 0.709\footnotesize{±0.039}& 0.745\footnotesize{±0.035}& \textcolor{blue}{\textbf{0.770}}\footnotesize{±0.014}& 0.682\footnotesize{±0.025}& \textcolor{red}{\textbf{0.777}}\footnotesize{±0.036} \\
          & Fuzzyball & 0.559\footnotesize{±0.008}& \textcolor{blue}{\textbf{0.617}}\footnotesize{±0.075}& 0.525\footnotesize{±0.043}& \textcolor{red}{\textbf{0.650}}\footnotesize{±0.064}& 0.525\footnotesize{±0.157}& 0.538\footnotesize{±0.092}& \textcolor{blue}{\textbf{0.734}}\footnotesize{±0.039}& 0.550\footnotesize{±0.082}& \textcolor{red}{\textbf{0.842}}\footnotesize{±0.026}& 0.677\footnotesize{±0.223}& 0.701\footnotesize{±0.093}\\
          & Nep & 0.566\footnotesize{±0.006}& \textcolor{blue}{\textbf{0.722}}\footnotesize{±0.023}& \textcolor{red}{\textbf{0.734}}\footnotesize{±0.038}& 0.710\footnotesize{±0.044}& 0.517\footnotesize{±0.059}& 0.717\footnotesize{±0.052}& \textcolor{red}{\textbf{0.810}}\footnotesize{±0.042}& 0.746\footnotesize{±0.060}& \textcolor{blue}{\textbf{0.771}}\footnotesize{±0.032}& 0.740\footnotesize{±0.052}& 0.750\footnotesize{±0.038} \\
          & Weft\_crack & 0.529\footnotesize{±0.006}& \textcolor{blue}{\textbf{0.586}}\footnotesize{±0.134}& 0.546\footnotesize{±0.114}& 0.582\footnotesize{±0.108}& 0.400\footnotesize{±0.029}& \textcolor{red}{\textbf{0.669}}\footnotesize{±0.045}& 0.599\footnotesize{±0.137}& \textcolor{blue}{\textbf{0.636}}\footnotesize{±0.051}& 0.618\footnotesize{±0.172}& 0.370\footnotesize{±0.037}& \textcolor{red}{\textbf{0.717}}\footnotesize{±0.072} \\\cline{2-13}
          & \textbf{Mean}  & 0.580\footnotesize{±0.004}& 0.646\footnotesize{±0.034}& 0.624\footnotesize{±0.024}& \textcolor{blue}{\textbf{0.674}}\footnotesize{±0.034}& 0.466\footnotesize{±0.030}& \textcolor{red}{\textbf{0.684}}\footnotesize{±0.033}& 0.683\footnotesize{±0.032}& 0.635\footnotesize{±0.043}& \textcolor{blue}{\textbf{0.724}}\footnotesize{±0.032}& 0.626\footnotesize{±0.041}& \textcolor{red}{\textbf{0.733}}\footnotesize{±0.009}  \\
    \hline
    \multirow{3}[0]{*}{\textbf{ELPV}} & Mono & 0.796\footnotesize{±0.002}& 0.634\footnotesize{±0.087}& \textcolor{blue}{\textbf{0.717}}\footnotesize{±0.025}& 0.563\footnotesize{±0.102}& 0.649\footnotesize{±0.027}& \textcolor{red}{\textbf{0.735}}\footnotesize{±0.031}& 0.599\footnotesize{±0.040}& 0.629\footnotesize{±0.072}& 0.569\footnotesize{±0.035}& \textcolor{red}{\textbf{0.756}}\footnotesize{±0.045}& \textcolor{blue}{\textbf{0.731}}\footnotesize{±0.021} \\
          & Poly & 0.679\footnotesize{±0.004}& 0.662\footnotesize{±0.050}& \textcolor{blue}{\textbf{0.665}}\footnotesize{±0.021}& \textcolor{blue}{\textbf{0.665}}\footnotesize{±0.173}& 0.483\footnotesize{±0.247}& \textcolor{red}{\textbf{0.671}}\footnotesize{±0.051}& \textcolor{red}{\textbf{0.804}}\footnotesize{±0.022}& 0.662\footnotesize{±0.042}& 0.796\footnotesize{±0.084}& 0.734\footnotesize{±0.078}& \textcolor{blue}{\textbf{0.800}}\footnotesize{±0.064} \\
          \cline{2-13}
          & \textbf{Mean}  & 0.737\footnotesize{±0.002}& 0.648\footnotesize{±0.057}& \textcolor{blue}{\textbf{0.691}}\footnotesize{±0.008}& 0.614\footnotesize{±0.048}& 0.566\footnotesize{±0.111}& \textcolor{red}{\textbf{0.703}}\footnotesize{±0.022}& 0.702\footnotesize{±0.023}& 0.646\footnotesize{±0.032}& 0.683\footnotesize{±0.047}&\textcolor{blue}{\textbf{ 0.745}}\footnotesize{±0.020}& \textcolor{red}{\textbf{0.766}}\footnotesize{±0.029} \\
    \hline
    \multirow{10}[0]{*}{\textbf{Mastcam}} 
          & Bedrock & 0.638\footnotesize{±0.007}& 0.495\footnotesize{±0.028}& 0.499\footnotesize{±0.056}& \textcolor{blue}{\textbf{0.636}}\footnotesize{±0.072}& 0.532\footnotesize{±0.036}& \textcolor{red}{\textbf{0.668}}\footnotesize{±0.012}& 0.550\footnotesize{±0.053}& 0.499\footnotesize{±0.098}& \textcolor{blue}{\textbf{0.636}}\footnotesize{±0.068}& 0.512\footnotesize{±0.062}& \textcolor{red}{\textbf{0.658}}\footnotesize{±0.021} \\
          & Broken-rock & 0.590\footnotesize{±0.007}& 0.533\footnotesize{±0.020}& 0.569\footnotesize{±0.025}& \textcolor{red}{\textbf{0.699}}\footnotesize{±0.058}& 0.544\footnotesize{±0.088}& \textcolor{blue}{\textbf{0.645}}\footnotesize{±0.053}& 0.547\footnotesize{±0.018}& 0.608\footnotesize{±0.085}& \textcolor{red}{\textbf{0.712}}\footnotesize{±0.052}& 0.651\footnotesize{±0.063}& \textcolor{blue}{\textbf{0.649}}\footnotesize{±0.047} \\
          & Drill-hole & 0.630\footnotesize{±0.006}& 0.555\footnotesize{±0.037}& 0.539\footnotesize{±0.077}& \textcolor{red}{\textbf{0.697}}\footnotesize{±0.074}& 0.636\footnotesize{±0.066}& \textcolor{blue}{\textbf{0.657}}\footnotesize{±0.070}& 0.583\footnotesize{±0.022}& 0.601\footnotesize{±0.009}& \textcolor{blue}{\textbf{0.682}}\footnotesize{±0.042}& 0.660\footnotesize{±0.002}& \textcolor{red}{\textbf{0.725}}\footnotesize{±0.005}\\
          & Drt  & 0.711\footnotesize{±0.005}& 0.529\footnotesize{±0.046}& 0.591\footnotesize{±0.042}& \textcolor{red}{\textbf{0.735}}\footnotesize{±0.020}& 0.624\footnotesize{±0.042}& \textcolor{blue}{\textbf{0.713}}\footnotesize{±0.053}& 0.621\footnotesize{±0.043}& 0.652\footnotesize{±0.024}& \textcolor{red}{\textbf{0.761}}\footnotesize{±0.062}& 0.616\footnotesize{±0.048}& \textcolor{blue}{\textbf{0.760}}\footnotesize{±0.033} \\
          & Dump-pile & 0.697\footnotesize{±0.007}& 0.521\footnotesize{±0.020}& 0.508\footnotesize{±0.021}& \textcolor{blue}{\textbf{0.682}}\footnotesize{±0.022}& 0.545\footnotesize{±0.127}& \textcolor{red}{\textbf{0.767}}\footnotesize{±0.043}& 0.705\footnotesize{±0.011}& 0.700\footnotesize{±0.070}& \textcolor{red}{\textbf{0.750}}\footnotesize{±0.037}& 0.696\footnotesize{±0.047}& \textcolor{blue}{\textbf{0.748}}\footnotesize{±0.066} \\
          & Float & 0.632\footnotesize{±0.007}& 0.502\footnotesize{±0.020}& 0.551\footnotesize{±0.030}& \textcolor{red}{\textbf{0.711}}\footnotesize{±0.041}& 0.530\footnotesize{±0.075}& \textcolor{blue}{\textbf{0.670}}\footnotesize{±0.065}& 0.615\footnotesize{±0.052}& \textcolor{blue}{\textbf{0.736}}\footnotesize{±0.041}& 0.718\footnotesize{±0.064}& 0.671\footnotesize{±0.032}& \textcolor{red}{\textbf{0.744}}\footnotesize{±0.073} \\
          & Meteorite & 0.634\footnotesize{±0.007}& 0.467\footnotesize{±0.049}& 0.462\footnotesize{±0.077}& \textcolor{red}{\textbf{0.669}}\footnotesize{±0.037}& 0.476\footnotesize{±0.014}& \textcolor{blue}{\textbf{0.637}}\footnotesize{±0.015}& 0.554\footnotesize{±0.021}& 0.568\footnotesize{±0.053}& \textcolor{blue}{\textbf{0.647}}\footnotesize{±0.030}& 0.473\footnotesize{±0.047}& \textcolor{red}{\textbf{0.716}}\footnotesize{±0.004} \\
          & Scuff & 0.638\footnotesize{±0.006}& 0.472\footnotesize{±0.031}& 0.508\footnotesize{±0.070}& \textcolor{red}{\textbf{0.679}}\footnotesize{±0.048}& 0.492\footnotesize{±0.037}& \textcolor{blue}{\textbf{0.549}}\footnotesize{±0.027}& 0.528\footnotesize{±0.034}& 0.575\footnotesize{±0.042}& \textcolor{red}{\textbf{0.676}}\footnotesize{±0.019}& 0.504\footnotesize{±0.052}& \textcolor{blue}{\textbf{0.636}}\footnotesize{±0.086} \\
          & Veins & 0.621\footnotesize{±0.007}& 0.527\footnotesize{±0.023}& 0.493\footnotesize{±0.052}& \textcolor{blue}{\textbf{0.688}}\footnotesize{±0.069}& 0.489\footnotesize{±0.028}& \textcolor{red}{\textbf{0.699}}\footnotesize{±0.045}& 0.589\footnotesize{±0.072}& 0.608\footnotesize{±0.044}& \textcolor{red}{\textbf{0.686}}\footnotesize{±0.053}& 0.510\footnotesize{±0.090}& \textcolor{blue}{\textbf{0.620}}\footnotesize{±0.036} \\
          \cline{2-13}
          & \textbf{Mean}  & 0.644\footnotesize{±0.003}& 0.511\footnotesize{±0.013}& 0.524\footnotesize{±0.013}& \textcolor{red}{\textbf{0.689}}\footnotesize{±0.037}& 0.541\footnotesize{±0.007}& \textcolor{blue}{\textbf{0.667}}\footnotesize{±0.012}& 0.588\footnotesize{±0.011}& 0.616\footnotesize{±0.021}& \textcolor{red}{\textbf{0.697}}\footnotesize{±0.014}& 0.588\footnotesize{±0.016}& \textcolor{blue}{\textbf{0.695}}\footnotesize{±0.004}  \\
    \hline
    \multirow{5}{*}{\textbf{Hyper-Kvasir}} & Barretts & 0.405\footnotesize{±0.003}& 0.672\footnotesize{±0.014}& \textcolor{blue}{\textbf{0.703}}\footnotesize{±0.040}& 0.382\footnotesize{±0.117}& 0.438\footnotesize{±0.111}& \textcolor{red}{\textbf{0.772}}\footnotesize{±0.019}& \textcolor{red}{\textbf{0.834}}\footnotesize{±0.012}& 0.764\footnotesize{±0.066}& 0.698\footnotesize{±0.037}& 0.540\footnotesize{±0.014}& \textcolor{blue}{\textbf{0.824}}\footnotesize{±0.006} \\
          & Barretts-short-seg & 0.404\footnotesize{±0.003}& \textcolor{blue}{\textbf{0.604}}\footnotesize{±0.048}& 0.538\footnotesize{±0.033}& 0.367\footnotesize{±0.050}& 0.532\footnotesize{±0.075}& \textcolor{red}{\textbf{0.674}}\footnotesize{±0.018}& 0.799\footnotesize{±0.036}& \textcolor{blue}{\textbf{0.810}}\footnotesize{±0.034}& 0.661\footnotesize{±0.034}& 0.480\footnotesize{±0.107}& \textcolor{red}{\textbf{0.835}}\footnotesize{±0.021} \\
          & Esophagitis-a & 0.435\footnotesize{±0.002}& \textcolor{blue}{\textbf{0.569}}\footnotesize{±0.051}& 0.536\footnotesize{±0.040}& 0.518\footnotesize{±0.063}& 0.491\footnotesize{±0.084}& \textcolor{red}{\textbf{0.778}}\footnotesize{±0.020}& \textcolor{blue}{\textbf{0.844}}\footnotesize{±0.014}& 0.815\footnotesize{±0.022}& 0.820\footnotesize{±0.034}& 0.646\footnotesize{±0.036}& \textcolor{red}{\textbf{0.881}}\footnotesize{±0.035} \\
          & Esophagitis-b-d &  0.367\footnotesize{±0.003}& \textcolor{blue}{\textbf{0.536}}\footnotesize{±0.033}& 0.505\footnotesize{±0.039}& 0.358\footnotesize{±0.039}& 0.457\footnotesize{±0.086}& \textcolor{red}{\textbf{0.577}}\footnotesize{±0.025}& \textcolor{blue}{\textbf{0.810}}\footnotesize{±0.015}& 0.754\footnotesize{±0.073}& 0.611\footnotesize{±0.017}& 0.621\footnotesize{±0.042}& \textcolor{red}{\textbf{0.837}}\footnotesize{±0.009} \\
          \cline{2-13}
          & \textbf{Mean}  & 0.403\footnotesize{±0.001}& \textcolor{blue}{\textbf{0.595}}\footnotesize{±0.023}& 0.571\footnotesize{±0.004}& 0.406\footnotesize{±0.01 8}& 0.480\footnotesize{±0.044}& \textcolor{red}{\textbf{0.700}}\footnotesize{±0.009}& \textcolor{blue}{\textbf{0.822}}\footnotesize{±0.019}& 0.786\footnotesize{±0.021}& 0.698\footnotesize{±0.021}& 0.571\footnotesize{±0.014}& \textcolor{red}{\textbf{0.844}}\footnotesize{±0.009} \\
    \hline
    \end{tabular}%
    }
  \label{tab:openset}%
  \vspace{-0.3cm}
\end{table*}%

The detection performance on six datasets applicable under the hard setting
is 
presented in Tab. \ref{tab:openset}.  

\textbf{Application Domain Perspective}. In both one-shot and ten-shot settings of the diverse application datasets, compared to the competing methods, our method is the best performer on most of the individual data subsets; at the dataset-level performance, our model achieves about 2\%-10\% mean AUC increase compared to the best contender on most of the six datasets, with close to the best performance on the other datasets. This shows substantially better generalizability of our model in detecting unseen anomaly classes than the other supervised detectors.

\textbf{Sample Efficiency}. Compared to one-shot scenarios to the ten-shot ones, our model, on average, has 5.5\% AUC decrease at the dataset level, 
which is better than that of the competing methods: DevNet (9.8\%), FLOS (7.1\%), SAOE (7.8\%), and MLEP (10\%). More impressively, our model trained with one anomaly example outperforms the ten-shot competing models by a large margin on many of the individual data subsets as well as the overall datasets.

\textbf{Comparison to Unsupervised Baseline}. Current supervised AD models are often biased towards the seen anomaly class and fail to generalize to unseen anomaly classes, performing less effective than the unsupervised baseline KDAD on most of the datasets. By contrast, our model has significantly improved generalizability and largely outperforms KDAD even under the one-shot scenarios. This generalizability is further supported by our preliminary results in cross-domain anomaly detection in \texttt{Appendix C.3}.

\subsection{Ablation Study} \label{sec:5.3.1}
\textbf{Importance of Each Abnormality Learning}.  
Tab. \ref{tab:ablation_image} shows the results of each abnormality learning in DRA using ten training anomaly examples.

\textbf{DRA1A} is the variant of DRA that only learns the known abnormality using the seen anomalies.
It performs comparably well to the two best contenders, DevNet and SAOE,
but limiting to the known abnormality leads to less effective results than the other variants, especially on the hard setting.

\textbf{DRA2A} builds upon DRA1A with the addition of pseudo abnormality learning.
Compared with DRA1A, the inclusion of the pseudo abnormality largely improves the performance on most datasets. 
On some datasets, such as Hyper-Kvasir on the hard setting, the performance of DRA2A drops, which may be due to the discrepancy between the pseudo and real anomalies. 

\textbf{DRA3Ar} and \textbf{DRA3An} extend DRA2A with one additional head. DRA3Ar attempts to learn the latent residual abnormality, without the support of the holistic normality representations that DRA3An is specifically designed to learn. DRA3Ar and DRA3An further largely improves DRA2A, but both of them are much less effective than our full model DRA. This demonstrates that $g_{r}$ and $g_{n}$ need to be fused to effectively learn the latent residual abnormality.

\begin{table}[tb]
  \centering
  \caption{Ablation study results of DRA and its variants. 
  `xA' denotes learning of `x' abnormalities.
  Best results are \textbf{highlighted}.}
  \vspace{-0.2cm}
  \scalebox{0.62}{
    \begin{tabular}{p{0.2cm}p{1.92cm}ccccc}
    \hline
    \multicolumn{2}{l}{\textbf{Module}}   & \multicolumn{1}{c}{\textbf{DRA1A}} & \multicolumn{1}{c}{\textbf{DRA2A}} &
    \multicolumn{1}{c}{\textbf{DRA3Ar}} &
    \multicolumn{1}{c}{\textbf{DRA3An}} & \multicolumn{1}{c}{\textbf{DRA}} \\ 
    \hline
    \multicolumn{2}{l}{$g_{s}$} & \multicolumn{1}{c}{$\checkmark$}  &\multicolumn{1}{c}{$\checkmark$} &\multicolumn{1}{c}{$\checkmark$} &\multicolumn{1}{c}{$\checkmark$} & \multicolumn{1}{c}{$\checkmark$} \\
    
    \multicolumn{2}{l}{$g_{p}$} & \multicolumn{1}{c}{} & \multicolumn{1}{c}{$\checkmark$} & \multicolumn{1}{c}{$\checkmark$} & \multicolumn{1}{c}{$\checkmark$}& \multicolumn{1}{c}{$\checkmark$} \\ 
    
    \multicolumn{2}{l}{$g_{r}$} & \multicolumn{1}{c}{} & \multicolumn{1}{c}{} & \multicolumn{1}{c}{$\checkmark$} & \multicolumn{1}{c}{} & \multicolumn{1}{c}{$\checkmark$} \\
    
    \multicolumn{2}{l}{$g_{n}$} & \multicolumn{1}{c}{} & \multicolumn{1}{c}{} &  \multicolumn{1}{c}{} & \multicolumn{1}{c}{$\checkmark$} & \multicolumn{1}{c}{$\checkmark$}\\
     \hline
        \multicolumn{7}{c}{\textbf{General Setting}}\\ 
    \hline
    \multicolumn{2}{l}{\textbf{MVTecAD}} & 0.938\footnotesize{±0.009}& 0.911\footnotesize{±0.012}& 0.927\footnotesize{±0.023}& 0.949\footnotesize{±0.006}& \textbf{0.959}\footnotesize{±0.003} \\
    \multicolumn{2}{l}{\textbf{AITEX}} & 0.881\footnotesize{±0.007}& \textbf{0.925}\footnotesize{±0.008}& 0.907\footnotesize{±0.014}& 0.898\footnotesize{±0.019}& 0.893\footnotesize{±0.017} \\
    \multicolumn{2}{l}{\textbf{SDD}} & 0.984\footnotesize{±0.013}& 0.984\footnotesize{±0.016}& 0.973\footnotesize{±0.021}& 0.988\footnotesize{±0.009}& \textbf{0.991}\footnotesize{±0.005} \\
    \multicolumn{2}{l}{\textbf{ELPV}} & 0.831\footnotesize{±0.011}& 0.794\footnotesize{±0.014}& 0.834\footnotesize{±0.039}& 0.823\footnotesize{±0.005}& \textbf{0.845}\footnotesize{±0.013} \\
    \multicolumn{2}{l}{\textbf{optical}} & 0.760\footnotesize{±0.038}& 0.946\footnotesize{±0.023}& 0.930\footnotesize{±0.002} & \textbf{0.965}\footnotesize{±0.007}& \textbf{0.965}\footnotesize{±0.006}  \\
     \multicolumn{2}{l}{\textbf{Mastcam}} & 0.756\footnotesize{±0.016}& 0.796\footnotesize{±0.008}& 0.813\footnotesize{±0.030}& 0.838\footnotesize{±0.016}& \textbf{0.848}\footnotesize{±0.008} \\
    \multicolumn{2}{l}{\textbf{BrainMRI}} & 0.965\footnotesize{±0.004}& 0.964\footnotesize{±0.007}& 0.958\footnotesize{±0.015}& 0.886\footnotesize{±0.030}& \textbf{0.970}\footnotesize{±0.003} \\
    \multicolumn{2}{l}{\textbf{HeadCT}} & 0.975\footnotesize{±0.003}& 0.974\footnotesize{±0.007}& 0.986\footnotesize{±0.007}& \textbf{0.988}\footnotesize{±0.006}& 0.972\footnotesize{±0.002} \\
    \multicolumn{2}{l}{\textbf{Hyper-Kvasir}} & 0.775\footnotesize{±0.026}& 0.790\footnotesize{±0.030}& 0.809\footnotesize{±0.026}& 0.725\footnotesize{±0.036}& \textbf{0.834}\footnotesize{±0.004} \\
    \hline
    \multicolumn{7}{c}{\textbf{Hard Setting}}\\ \hline
    \multirow{6}{*}{\rotatebox{90}{\textbf{Carpet}}} & Color & 0.739\footnotesize{±0.007}& 0.671\footnotesize{±0.167}& 0.847\footnotesize{±0.045}& 0.848\footnotesize{±0.062}& \textbf{0.886}\footnotesize{±0.042} \\
          & Cut & 0.731\footnotesize{±0.055}& 0.880\footnotesize{±0.021}& 0.763\footnotesize{±0.176}& 0.885\footnotesize{±0.080}& \textbf{0.922}\footnotesize{±0.038}\\
          & Hole & 0.735\footnotesize{±0.077}& 0.733\footnotesize{±0.116}& 0.903\footnotesize{±0.049}& 0.903\footnotesize{±0.044}& \textbf{0.947}\footnotesize{±0.016}\\
          & Metal & 0.768\footnotesize{±0.035}& 0.860\footnotesize{±0.044}& 0.896\footnotesize{±0.025}& 0.868\footnotesize{±0.078}& \textbf{0.933}\footnotesize{±0.022}\\
          & Thread & 0.970\footnotesize{±0.016}& 0.978\footnotesize{±0.005}& 0.985\footnotesize{±0.007}& \textbf{0.992}\footnotesize{±0.006}& 0.989\footnotesize{±0.004} \\
          \cline{2-7}
          & \textbf{Mean} & 0.788\footnotesize{±0.025}& 0.824\footnotesize{±0.045}& 0.879\footnotesize{±0.047}& 0.899\footnotesize{±0.014}& \textbf{0.935}\footnotesize{±0.013}\\
    \hline
    \multirow{7}[0]{*}{\rotatebox{90}{\textbf{AITEX}}} & Broken\_end & 0.638\footnotesize{±0.019}& 0.738\footnotesize{±0.142}& \textbf{0.744}\footnotesize{±0.114}& 0.640\footnotesize{±0.128}& 0.693\footnotesize{±0.099} \\
          & Broken\_pick & 0.651\footnotesize{±0.037}& 0.714\footnotesize{±0.039}& 0.675\footnotesize{±0.047}& 0.725\footnotesize{±0.104}& \textbf{0.760}\footnotesize{±0.037} \\
          & Cut\_selvage & 0.710\footnotesize{±0.019}& 0.724\footnotesize{±0.048}& 0.766\footnotesize{±0.035}& 0.702\footnotesize{±0.032}& \textbf{0.777}\footnotesize{±0.036}\\
          & Fuzzyball & \textbf{0.714}\footnotesize{±0.019}& 0.676\footnotesize{±0.038}& 0.654\footnotesize{±0.102}& 0.631\footnotesize{±0.014}& 0.701\footnotesize{±0.093}\\
          & Nep & 0.775\footnotesize{±0.027}& 0.745\footnotesize{±0.036}& 0.759\footnotesize{±0.047}& \textbf{0.784}\footnotesize{±0.034}& 0.750\footnotesize{±0.038} \\
          & Weft\_crack & 0.633\footnotesize{±0.073}& 0.636\footnotesize{±0.079}& \textbf{0.768}\footnotesize{±0.077}& 0.735\footnotesize{±0.110}& 0.717\footnotesize{±0.072}\\\cline{2-7}
          & \textbf{Mean} & 0.687\footnotesize{±0.018}& 0.706\footnotesize{±0.041}& 0.728\footnotesize{±0.027}& 0.703\footnotesize{±0.054}& \textbf{0.733}\footnotesize{±0.009}   \\
    \hline
    \multirow{3}[0]{*}{\rotatebox{90}{\textbf{ELPV}}} & Mono & 0.631\footnotesize{±0.042}& 0.655\footnotesize{±0.034}& 0.684\footnotesize{±0.050}& 0.650\footnotesize{±0.034}& \textbf{0.731}\footnotesize{±0.021}\\
          & Poly & 0.761\footnotesize{±0.033}& 0.823\footnotesize{±0.016}&  0.808\footnotesize{±0.067}& \textbf{0.837}\footnotesize{±0.045}& 0.800\footnotesize{±0.064} \\
          \cline{2-7}
          & \textbf{Mean} & 0.696\footnotesize{±0.005}& 0.739\footnotesize{±0.025}& 0.746\footnotesize{±0.048}& 0.744\footnotesize{±0.039}& \textbf{0.766}\footnotesize{±0.029} \\
    \hline
    \multirow{5}{*}{\rotatebox{90}{\textbf{Hyper-Kvasir}}} & Barretts & \textbf{0.833}\footnotesize{±0.028}& 0.731\footnotesize{±0.022}& 0.778\footnotesize{±0.025}& 0.819\footnotesize{±0.030}& 0.824\footnotesize{±0.006}\\
          & 
          B.-short-seg& 0.810\footnotesize{±0.050}& 0.741\footnotesize{±0.052}& 0.688\footnotesize{±0.076}& 0.825\footnotesize{±0.038}& \textbf{0.835}\footnotesize{±0.021}\\
          &  
          Esophagitis-a& 0.840\footnotesize{±0.030}& 0.816\footnotesize{±0.045}& 0.789\footnotesize{±0.060}& \textbf{0.889}\footnotesize{±0.010}& 0.881\footnotesize{±0.035}\\
          &  
          E.-b-d& 0.741\footnotesize{±0.031}& 0.633\footnotesize{±0.046}& 0.652\footnotesize{±0.069}& 0.805\footnotesize{±0.006}& \textbf{0.837}\footnotesize{±0.009}\\
          \cline{2-7}
          & \textbf{Mean}  & 0.806\footnotesize{±0.014}& 0.730\footnotesize{±0.040}& 0.727\footnotesize{±0.032}& 0.835\footnotesize{±0.007}& \textbf{0.844}\footnotesize{±0.009}\\
    \hline
    \end{tabular}%
    }
    \vspace{-0.2cm}
  \label{tab:ablation_image}%
\end{table}%

\begin{figure}[t] 
    \centering
    \begin{subfigure}[b]{0.47\linewidth}
         \centering    
            \includegraphics[width=1.\linewidth]{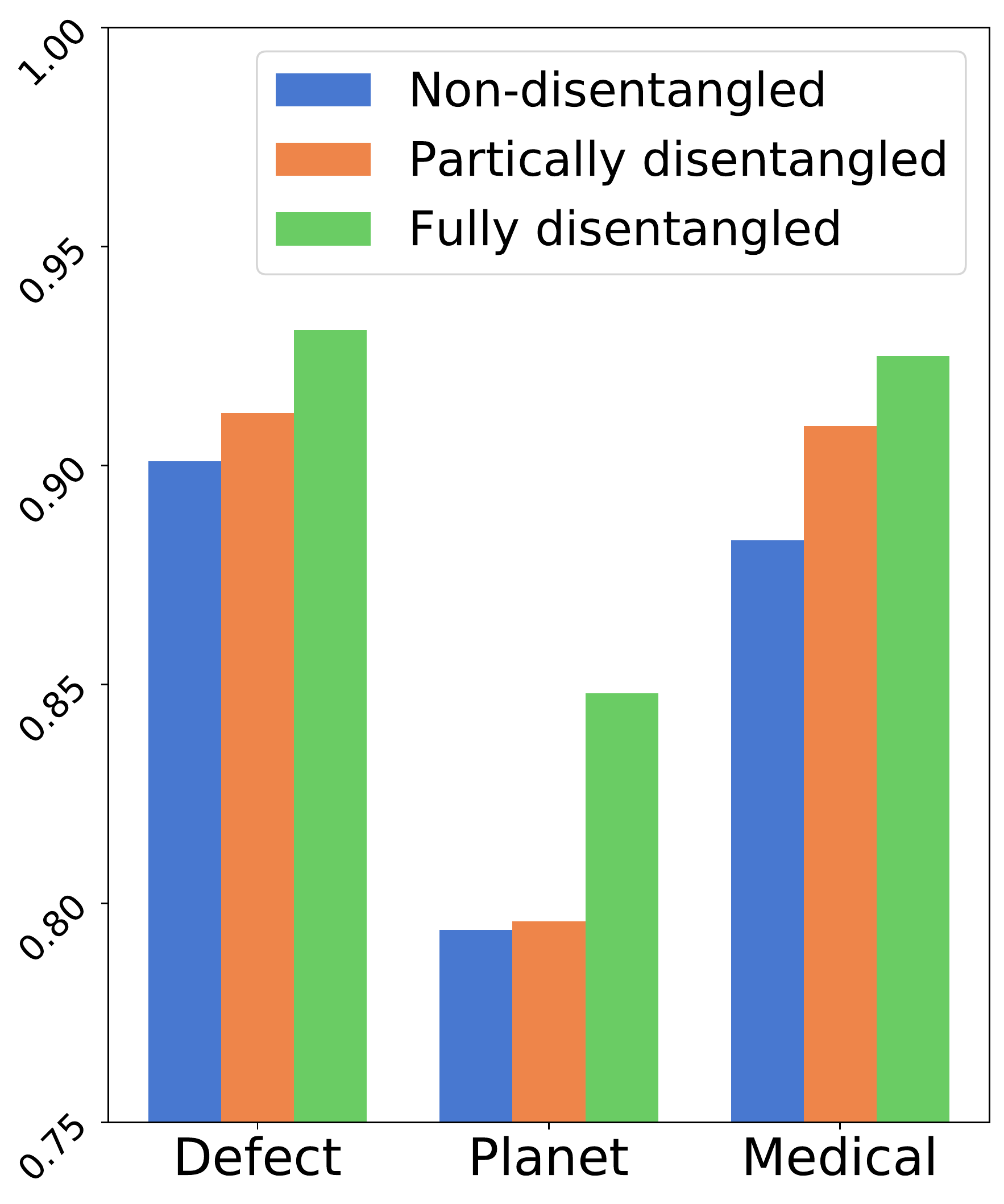}
            \label{fig:ref_sen}
    \end{subfigure}
        \begin{subfigure}[b]{0.47\linewidth}
         \centering     \includegraphics[width=1.\linewidth]{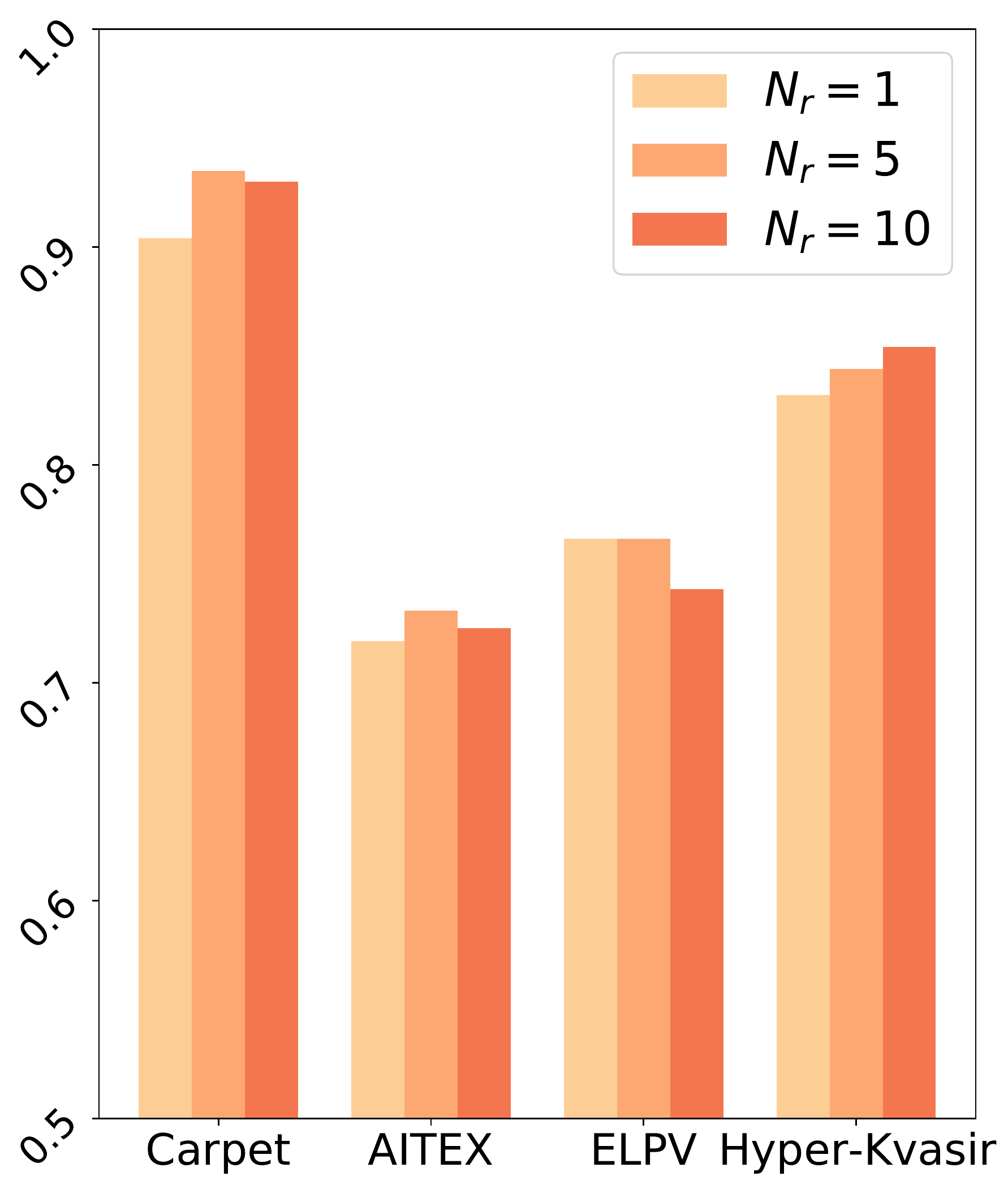}
         \label{fig:disentangle}
    \end{subfigure}
    \vspace{-0.55cm}
    \caption{(\textbf{Left}) Disentangled vs. non-disentangled abnormality learning. The results are averaged over the datasets in each domain. (\textbf{Right}) AUC results of DRA using different reference set sizes ($\mathit{N}_{r}$). Each result is averaged over all data subsets per dataset.}
    \label{fig:combined}
    \vspace{-0.5cm}
\end{figure}

\textbf{Importance of Disentangled Abnormalities}. Fig. \ref{fig:combined} (Left) shows the results of non-disentangled, partially, and fully disentangled abnormality learning under the general setting, where the non-disentangled method is multi-class classification with normal samples, seen and pseudo anomaly classes, the partially disentangled method is the variant of DRA that learns disentangled seen and pseudo abnormalities only, and the fully disentangled method is our model DRA. The results show that disentangled abnormality learning helps largely improve the detection performance across three application domains.

\begin{table}[bt]
  \centering
  \caption{AUC results w.r.t. methods to create pseudo anomalies.}
  \vspace{-0.2cm}
  \scalebox{0.62}{
    \begin{tabular}{p{0.2cm}p{2.2cm}|ccc|cc}
    \hline
    \multirow{2}{*}{\textbf{Anomaly Category}}      & \multirow{2}{*}{\textbf{}} &  \multicolumn{3}{c|}{\textbf{Augmentation}} & \multicolumn{2}{c}{\textbf{External}}\\
    \cline{3-7} 
    && \multicolumn{1}{c}{\textbf{CP-Scar}} & \multicolumn{1}{c}{\textbf{CP-Mix}} & 
    \multicolumn{1}{c|}{\textbf{CutMix}} & 
    \multicolumn{1}{c}{\textbf{MVTec AD}} & \multicolumn{1}{c}{\textbf{LAG}}\\
    \hline
    \multirow{6}{*}{\rotatebox{90}{\textbf{Carpet}}} & Color & 0.743\footnotesize{±0.142}& \textbf{0.967}\footnotesize{±0.048}& 0.886\footnotesize{±0.042}& 0.615\footnotesize{±0.028}& 0.711\footnotesize{±0.041}  \\
          & Cut & 0.853\footnotesize{±0.098}& 0.862\footnotesize{±0.072}& \textbf{0.922}\footnotesize{±0.038}& 0.688\footnotesize{±0.019}& 0.721\footnotesize{±0.021}\\
          & Hole & 0.809\footnotesize{±0.033}& \textbf{0.955}\footnotesize{±0.024}& 0.947\footnotesize{±0.016}& 0.712\footnotesize{±0.015}& 0.823\footnotesize{±0.020}\\
          & Metal & 0.858\footnotesize{±0.197}& 0.840\footnotesize{±0.096}& \textbf{0.933}\footnotesize{±0.022}& 0.764\footnotesize{±0.039}& 0.670\footnotesize{±0.037} \\
          & Thread & 0.987\footnotesize{±0.013}& 0.988\footnotesize{±0.011}& \textbf{0.989}\footnotesize{±0.004}& 0.966\footnotesize{±0.003}& 0.968\footnotesize{±0.005} \\
          \cline{2-7}
          & \textbf{Mean}  & 0.850\footnotesize{±0.070}& 0.922\footnotesize{±0.012}& \textbf{0.935}\footnotesize{±0.013}& 0.749\footnotesize{±0.006}& 0.779\footnotesize{±0.017}\\
    \hline
    \multirow{7}[0]{*}{\rotatebox{90}{\textbf{AITEX}}} & Broken\_end & 0.584\footnotesize{±0.127}& 0.750\footnotesize{±0.115}& 0.693\footnotesize{±0.099}& \textbf{0.793}\footnotesize{±0.043}& 0.722\footnotesize{±0.072}\\
          & Broken\_pick & 0.616\footnotesize{±0.111}& 0.671\footnotesize{±0.082}& \textbf{0.760}\footnotesize{±0.037}& 0.603\footnotesize{±0.017}& 0.584\footnotesize{±0.034}\\
          & Cut\_selvage & 0.676\footnotesize{±0.032}& 0.653\footnotesize{±0.091}& \textbf{0.777}\footnotesize{±0.036}& 0.690\footnotesize{±0.013}& 0.683\footnotesize{±0.035}\\
          & Fuzzyball & 0.639\footnotesize{±0.056}& 0.582\footnotesize{±0.067}& 0.701\footnotesize{±0.093}& \textbf{0.743}\footnotesize{±0.053}& 0.588\footnotesize{±0.112} \\
          & Nep & 0.679\footnotesize{±0.060}& 0.706\footnotesize{±0.096}& 0.750\footnotesize{±0.038}& \textbf{0.774}\footnotesize{±0.029}& 0.739\footnotesize{±0.012} \\
          & Weft\_crack & 0.470\footnotesize{±0.209}& 0.507\footnotesize{±0.293}& \textbf{0.717}\footnotesize{±0.072}& 0.671\footnotesize{±0.031}& 0.480\footnotesize{±0.140} \\
          \cline{2-7}
          & \textbf{Mean}  & 0.611\footnotesize{±0.064}& 0.645\footnotesize{±0.070}& \textbf{0.733}\footnotesize{±0.009}& 0.712\footnotesize{±0.010}& 0.633\footnotesize{±0.049} \\
    \hline
    \multirow{3}[0]{*}{\rotatebox{90}{\textbf{ELPV}}} & Mono & 0.665\footnotesize{±0.098}& 0.622\footnotesize{±0.067}& \textbf{0.731}\footnotesize{±0.021}& 0.543\footnotesize{±0.064}& 0.544\footnotesize{±0.041}\\
          & Poly & 0.755\footnotesize{±0.006}& 0.807\footnotesize{±0.085}& 0.800\footnotesize{±0.064}& 0.749\footnotesize{±0.052}& \textbf{0.808}\footnotesize{±0.056}\\
          \cline{2-7}
          & \textbf{Mean} & 0.710\footnotesize{±0.046}& 0.715\footnotesize{±0.076}& \textbf{0.766}\footnotesize{±0.029}& 0.646\footnotesize{±0.042}& 0.676\footnotesize{±0.031} \\
    \hline
    \multirow{5}{*}{\rotatebox{90}{\textbf{Hyper-Kvasir}}} & Barretts & 0.832\footnotesize{±0.016}& 0.735\footnotesize{±0.028}& 0.761\footnotesize{±0.043}& \textbf{0.834}\footnotesize{±0.024}& 0.824\footnotesize{±0.006}\\
          & B.-short-seg & 0.827\footnotesize{±0.054}& 0.719\footnotesize{±0.049}& 0.695\footnotesize{±0.030}& \textbf{0.839}\footnotesize{±0.038}& 0.835\footnotesize{±0.021}\\
          & Esophagitis-a & 0.832\footnotesize{±0.024}& 0.751\footnotesize{±0.023}& 0.763\footnotesize{±0.070}& 0.811\footnotesize{±0.031}& \textbf{0.881}\footnotesize{±0.035} \\
          & E.-b-d & 0.805\footnotesize{±0.035}& 0.749\footnotesize{±0.060}& 0.782\footnotesize{±0.028}& \textbf{0.847}\footnotesize{±0.017}& 0.837\footnotesize{±0.009} \\
          \cline{2-7}
          & \textbf{Mean}  & 0.824\footnotesize{±0.020}& 0.739\footnotesize{±0.007}& 0.751\footnotesize{±0.021}& 0.833\footnotesize{±0.023}& \textbf{0.844}\footnotesize{±0.009} \\
    \hline
    \end{tabular}%
    }
  \label{tab:dummy}%
  \vspace{-0.5cm}
\end{table}%

\subsection{Sensitivity Analysis} \label{sec:sensitivity}

\textbf{Sensitivity w.r.t.\  the Source of Pseudo Anomalies} 
There are diverse ways to create pseudo anomalies, as shown in previous studies \cite{li2021cutpaste,reiss2021panda,tack2020csi,hendrycks2018deep} that focus on anomaly-free training data. 
We instead investigate the effect of these methods under our open-set supervised AD model DRA. We evaluate three data augmentation methods, including CutMix \cite{yun2019cutmix}, CutPaste-Scar (CP-Scar) \cite{li2021cutpaste} and CutPaste-Mix (CP-Mix) that utilizes both CutMix and CP-Scar, and two outlier exposure methods that respectively use samples from MVTec AD \cite{Bergmann_2019_CVPR} and medical dataset LAG \cite{Li_2019_CVPR} as the pseudo anomalies.
When using MVTec AD, we remove the classes that overlap with the training/test data; LAG does not have any overlapping with our datasets. Since pseudo anomalies are used mainly to enhance the generalization to unseen anomalies, we focus on the four hard setting datasets in our ablation study in Tab. \ref{tab:ablation_image}.

The results are shown in Tab.~\ref{tab:dummy}, from which it is clear that the data augmentation-based pseudo anomaly creation methods are generally more stable and much better than the external data-based methods on non-medical datasets. 
On the other hand, the external data method is more effective on medical datasets, since the augmentation methods often fail to properly simulate the lesions. The LAG dataset provides more application-relevant features and enables DRA to achieve the best results on Hyper-Kvasir.

\textbf{Sensitivity w.r.t. the Reference Size in Latent Residual Abnormality Learning}. Our latent residual abnormality learning head requires to sample a fixed number $\mathit{N}_{r}$ of normal training images as reference data. We evaluate the sensitivity of our method using different $\mathit{N}_{r}$ and report the AUC results in Fig. \ref{fig:combined} (Right). 
Using one reference image is generally sufficient to learn the residual anomalies. 
Increasing the reference size to five helps further improve the detection performance, but increasing the size to ten is not consistently helpful. $\mathit{N}_{r}=5$ is generally recommended, which is the default setting in DRA in all our experiments.
\vspace{-0.1cm}

\section{Conclusions and Discussions}

This paper proposes the framework of learning disentangled representations of abnormalities illustrated by seen anomalies, pseudo anomalies, and latent residual-based anomalies, and introduces the DRA model to effectively detect both seen and unseen anomalies. Our comprehensive results in Tabs. \ref{tab:randomanomalies} and \ref{tab:openset} justify that these three disentangled abnormality representations can complement each other in detecting the largely varying anomalies, substantially outperforming five SotA unsupervised and supervised anomaly detectors by a large margin, especially on the challenging cases, \eg, having only one training anomaly example, or detecting unseen anomalies. 
 
 The studied problem is largely under-explored, but it is very important in many relevant real-world applications. As shown by the results in Tabs. \ref{tab:randomanomalies} and \ref{tab:openset}, there are still a number of major challenges requiring further investigation, \eg, generalization from smaller anomaly examples from fewer classes, of which our model and comprehensive results provide a good baseline and extensive benchmark results.

{\small
\bibliographystyle{ieee_fullname}
\bibliography{egbib}

\begin{thebibliography}{10}\itemsep=-1pt

\bibitem{bendale2016towards}
Abhijit Bendale and Terrance~E Boult.
\newblock Towards open set deep networks.
\newblock In {\em Proc. IEEE Conf. Comp. Vis. Patt. Recogn.}, pages 1563--1572,
  2016.

\bibitem{bergman2020goad}
Liron Bergman and Yedid Hoshen.
\newblock Classification-based anomaly detection for general data.
\newblock In {\em Proc. Int. Conf. Learn. Representations}, 2020.

\bibitem{Bergmann_2019_CVPR}
Paul Bergmann, Michael Fauser, David Sattlegger, and Carsten Steger.
\newblock Mvtec ad -- a comprehensive real-world dataset for unsupervised
  anomaly detection.
\newblock In {\em Proc. IEEE Conf. Comp. Vis. Patt. Recogn.}, June 2019.

\bibitem{Bergmann_2020_CVPR}
Paul Bergmann, Michael Fauser, David Sattlegger, and Carsten Steger.
\newblock Uninformed students: Student-teacher anomaly detection with
  discriminative latent embeddings.
\newblock In {\em Proc. IEEE Conf. Comp. Vis. Patt. Recogn.}, June 2020.

\bibitem{borgli2020hyperkvasir}
Hanna Borgli, Vajira Thambawita, Pia~H Smedsrud, Steven Hicks, Debesh Jha,
  Sigrun~L Eskeland, Kristin~Ranheim Randel, Konstantin Pogorelov, Mathias Lux,
  Duc Tien~Dang Nguyen, et~al.
\newblock Hyperkvasir, a comprehensive multi-class image and video dataset for
  gastrointestinal endoscopy.
\newblock {\em Scientific Data}, 7(1):1--14, 2020.

\bibitem{branco2016survey}
Paula Branco, Lu{\'\i}s Torgo, and Rita Ribeiro.
\newblock A survey of predictive modeling on imbalanced domains.
\newblock {\em ACM Computing Surveys}, 49(2):1--50, 2016.

\bibitem{chalapathy2018anomaly}
Raghavendra Chalapathy, Aditya~Krishna Menon, and Sanjay Chawla.
\newblock Anomaly detection using one-class neural networks.
\newblock {\em arXiv preprint arXiv:1802.06360}, 2018.

\bibitem{chen2021deep}
Yuanhong Chen, Yu Tian, Guansong Pang, and Gustavo Carneiro.
\newblock Deep one-class classification via interpolated gaussian descriptor.
\newblock In {\em Proc. {AAAI} Conf. Artificial Intell.}, 2022.

\bibitem{deitsch2019elpv}
Sergiu Deitsch, Vincent Christlein, Stephan Berger, Claudia Buerhop-Lutz,
  Andreas Maier, Florian Gallwitz, and Christian Riess.
\newblock Automatic classification of defective photovoltaic module cells in
  electroluminescence images.
\newblock {\em Solar Energy}, 185:455--468, 2019.

\bibitem{di2021pixel}
Giancarlo Di~Biase, Hermann Blum, Roland Siegwart, and Cesar Cadena.
\newblock Pixel-wise anomaly detection in complex driving scenes.
\newblock In {\em Proc. IEEE Conf. Comp. Vis. Patt. Recogn.}, pages
  16918--16927, 2021.

\bibitem{georgescu2021anomaly}
Mariana-Iuliana Georgescu, Antonio Barbalau, Radu~Tudor Ionescu, Fahad~Shahbaz
  Khan, Marius Popescu, and Mubarak Shah.
\newblock Anomaly detection in video via self-supervised and multi-task
  learning.
\newblock In {\em Proc. IEEE Conf. Comp. Vis. Patt. Recogn.}, pages
  12742--12752, 2021.

\bibitem{golan2018deep}
Izhak Golan and Ran El-Yaniv.
\newblock Deep anomaly detection using geometric transformations.
\newblock In {\em Proc. Advances in Neural Inf. Process. Syst.}, pages
  9758--9769, 2018.

\bibitem{gong2019memorizing}
Dong Gong, Lingqiao Liu, Vuong Le, Budhaditya Saha, Moussa~Reda Mansour, Svetha
  Venkatesh, and Anton van~den Hengel.
\newblock Memorizing normality to detect anomaly: Memory-augmented deep
  autoencoder for unsupervised anomaly detection.
\newblock In {\em Proc. IEEE Int. Conf. Comp. Vis.}, pages 1705--1714, 2019.

\bibitem{gornitz2013toward}
Nico G{\"o}rnitz, Marius Kloft, Konrad Rieck, and Ulf Brefeld.
\newblock Toward supervised anomaly detection.
\newblock {\em J. Artificial Intelligence Research}, 46:235--262, 2013.

\bibitem{hadsell2006contrastloss}
R. Hadsell, S. Chopra, and Y. LeCun.
\newblock Dimensionality reduction by learning an invariant mapping.
\newblock In {\em Proc. IEEE Conf. Comp. Vis. Patt. Recogn.}, volume~2, pages
  1735--1742, 2006.

\bibitem{he2009imbalance}
Haibo He and Edwardo~A Garcia.
\newblock Learning from imbalanced data.
\newblock {\em IEEE T. Knowledge and Data Engineering}, 21(9):1263--1284, 2009.

\bibitem{hendrycks17baseline}
Dan Hendrycks and Kevin Gimpel.
\newblock A baseline for detecting misclassified and out-of-distribution
  examples in neural networks.
\newblock In {\em Proc. Int. Conf. Learn. Representations}, 2017.

\bibitem{hendrycks2018deep}
Dan Hendrycks, Mantas Mazeika, and Thomas Dietterich.
\newblock Deep anomaly detection with outlier exposure.
\newblock In {\em Proc. Int. Conf. Learn. Representations}, 2019.

\bibitem{Hou_2021_ICCV}
Jinlei Hou, Yingying Zhang, Qiaoyong Zhong, Di Xie, Shiliang Pu, and Hong Zhou.
\newblock Divide-and-assemble: Learning block-wise memory for unsupervised
  anomaly detection.
\newblock In {\em Proc. IEEE Int. Conf. Comp. Vis.}, pages 8791--8800, October
  2021.

\bibitem{huang2021mos}
Rui Huang and Yixuan Li.
\newblock Mos: Towards scaling out-of-distribution detection for large semantic
  space.
\newblock In {\em Proc. IEEE Conf. Comp. Vis. Patt. Recogn.}, pages 8710--8719,
  2021.

\bibitem{tudor2017unmasking}
Radu~Tudor Ionescu, Sorina Smeureanu, Bogdan Alexe, and Marius Popescu.
\newblock Unmasking the abnormal events in video.
\newblock In {\em Proc. IEEE Int. Conf. Comp. Vis.}, pages 2895--2903, 2017.

\bibitem{kerner2020comparison}
Hannah~R Kerner, Kiri~L Wagstaff, Brian~D Bue, Danika~F Wellington, Samantha
  Jacob, Paul Horton, James~F Bell, Chiman Kwan, and Heni~Ben Amor.
\newblock Comparison of novelty detection methods for multispectral images in
  rover-based planetary exploration missions.
\newblock {\em Data Mining and Knowledge Discovery}, 34(6):1642--1675, 2020.

\bibitem{khakzad2015grayswan}
Nima Khakzad, Faisal Khan, and Paul Amyotte.
\newblock Major accidents (gray swans) likelihood modeling using accident
  precursors and approximate reasoning.
\newblock {\em Risk analysis}, 35(7):1336--1347, 2015.

\bibitem{krizhevsky2009learning}
Alex Krizhevsky, Geoffrey Hinton, et~al.
\newblock Learning multiple layers of features from tiny images.
\newblock 2009.

\bibitem{lecun1998gradient}
Yann LeCun, L{\'e}on Bottou, Yoshua Bengio, and Patrick Haffner.
\newblock Gradient-based learning applied to document recognition.
\newblock {\em Proceedings of the IEEE}, 86(11):2278--2324, 1998.

\bibitem{li2021cutpaste}
Chun-Liang Li, Kihyuk Sohn, Jinsung Yoon, and Tomas Pfister.
\newblock Cutpaste: Self-supervised learning for anomaly detection and
  localization.
\newblock In {\em Proc. IEEE Conf. Comp. Vis. Patt. Recogn.}, pages 9664--9674,
  2021.

\bibitem{Li_2019_CVPR}
Liu Li, Mai Xu, Xiaofei Wang, Lai Jiang, and Hanruo Liu.
\newblock Attention based glaucoma detection: A large-scale database and cnn
  model.
\newblock In {\em Proc. IEEE Conf. Comp. Vis. Patt. Recogn.}, June 2019.

\bibitem{lin2017focalloss}
Tsung-Yi Lin, Priya Goyal, Ross Girshick, Kaiming He, and Piotr Doll{\'a}r.
\newblock Focal loss for dense object detection.
\newblock In {\em Proc. IEEE Int. Conf. Comp. Vis.}, pages 2980--2988, 2017.

\bibitem{Lin_2021_CVPR}
Ziqian Lin, Sreya~Dutta Roy, and Yixuan Li.
\newblock Mood: Multi-level out-of-distribution detection.
\newblock In {\em Proc. IEEE Conf. Comp. Vis. Patt. Recogn.}, pages
  15313--15323, June 2021.

\bibitem{liu2019margin}
Wen Liu, Weixin Luo, Zhengxin Li, Peilin Zhao, Shenghua Gao, et~al.
\newblock Margin learning embedded prediction for video anomaly detection with
  a few anomalies.
\newblock In {\em Proc. Int. Joint Conf. Artificial Intell.}, pages 3023--3030,
  2019.

\bibitem{liu2019large}
Ziwei Liu, Zhongqi Miao, Xiaohang Zhan, Jiayun Wang, Boqing Gong, and Stella~X
  Yu.
\newblock Large-scale long-tailed recognition in an open world.
\newblock In {\em Proc. IEEE Conf. Comp. Vis. Patt. Recogn.}, pages 2537--2546,
  2019.

\bibitem{liznerski2021explainable}
Philipp Liznerski, Lukas Ruff, Robert~A. Vandermeulen, Billy~Joe Franks, Marius
  Kloft, and Klaus-Robert M{\"u}ller.
\newblock Explainable deep one-class classification.
\newblock In {\em Proc. Int. Conf. Learn. Representations}, 2021.

\bibitem{Markovitz_2020_CVPR}
Amir Markovitz, Gilad Sharir, Itamar Friedman, Lihi Zelnik-Manor, and Shai
  Avidan.
\newblock Graph embedded pose clustering for anomaly detection.
\newblock In {\em Proc. IEEE Conf. Comp. Vis. Patt. Recogn.}, June 2020.

\bibitem{pang2018learning}
Guansong Pang, Longbing Cao, Ling Chen, and Huan Liu.
\newblock Learning representations of ultrahigh-dimensional data for random
  distance-based outlier detection.
\newblock In {\em Proc. {ACM SIGKDD} Int. Conf. Knowledge Discovery \& Data
  Mining}, pages 2041--2050, 2018.

\bibitem{pang2021explainable}
Guansong Pang, Choubo Ding, Chunhua Shen, and Anton van~den Hengel.
\newblock Explainable deep few-shot anomaly detection with deviation networks.
\newblock {\em arXiv preprint arXiv:2108.00462}, 2021.

\bibitem{pang2021deep}
Guansong Pang, Chunhua Shen, Longbing Cao, and Anton Van~Den Hengel.
\newblock Deep learning for anomaly detection: A review.
\newblock {\em ACM Computing Surveys}, 54(2):1--38, 2021.

\bibitem{pang2019deep}
Guansong Pang, Chunhua Shen, and Anton van~den Hengel.
\newblock Deep anomaly detection with deviation networks.
\newblock In {\em Proc. {ACM SIGKDD} Int. Conf. Knowledge Discovery \& Data
  Mining}, pages 353--362, 2019.

\bibitem{pang2021toward}
Guansong Pang, Anton van~den Hengel, Chunhua Shen, and Longbing Cao.
\newblock Toward deep supervised anomaly detection: Reinforcement learning from
  partially labeled anomaly data.
\newblock In {\em Proc. {ACM SIGKDD} Int. Conf. Knowledge Discovery \& Data
  Mining}, pages 1298--1308, 2021.

\bibitem{Park_2020_CVPR}
Hyunjong Park, Jongyoun Noh, and Bumsub Ham.
\newblock Learning memory-guided normality for anomaly detection.
\newblock In {\em Proc. IEEE Conf. Comp. Vis. Patt. Recogn.}, June 2020.

\bibitem{Perera_2019_CVPR}
Pramuditha Perera, Ramesh Nallapati, and Bing Xiang.
\newblock Ocgan: One-class novelty detection using gans with constrained latent
  representations.
\newblock In {\em Proc. IEEE Conf. Comp. Vis. Patt. Recogn.}, June 2019.

\bibitem{perera2019learning}
Pramuditha Perera and Vishal~M Patel.
\newblock Learning deep features for one-class classification.
\newblock {\em IEEE Transactions on Image Processing}, 28(11):5450--5463, 2019.

\bibitem{reiss2021panda}
Tal Reiss, Niv Cohen, Liron Bergman, and Yedid Hoshen.
\newblock Panda: Adapting pretrained features for anomaly detection and
  segmentation.
\newblock In {\em Proc. IEEE Conf. Comp. Vis. Patt. Recogn.}, pages 2806--2814,
  2021.

\bibitem{ren2019likelihood}
Jie Ren, Peter~J Liu, Emily Fertig, Jasper Snoek, Ryan Poplin, Mark~A DePristo,
  Joshua~V Dillon, and Balaji Lakshminarayanan.
\newblock Likelihood ratios for out-of-distribution detection.
\newblock In {\em Proc. Advances in Neural Inf. Process. Syst.}, 2019.

\bibitem{ruff2018deep}
Lukas Ruff, Robert Vandermeulen, Nico Goernitz, Lucas Deecke, Shoaib~Ahmed
  Siddiqui, Alexander Binder, Emmanuel M{\"u}ller, and Marius Kloft.
\newblock Deep one-class classification.
\newblock In {\em Proc. Int. Conf. Mach. Learn.}, pages 4393--4402, 2018.

\bibitem{ruff2019deep}
Lukas Ruff, Robert~A Vandermeulen, Nico G{\"o}rnitz, Alexander Binder, Emmanuel
  M{\"u}ller, Klaus-Robert M{\"u}ller, and Marius Kloft.
\newblock Deep semi-supervised anomaly detection.
\newblock In {\em Proc. Int. Conf. Learn. Representations}, 2020.

\bibitem{sabokrou2018adversarially}
Mohammad Sabokrou, Mohammad Khalooei, Mahmood Fathy, and Ehsan Adeli.
\newblock Adversarially learned one-class classifier for novelty detection.
\newblock In {\em Proc. IEEE Conf. Comp. Vis. Patt. Recogn.}, pages 3379--3388,
  2018.

\bibitem{salehi2021multiresolution}
Mohammadreza Salehi, Niousha Sadjadi, Soroosh Baselizadeh, Mohammad~Hossein
  Rohban, and Hamid~R Rabiee.
\newblock Multiresolution knowledge distillation for anomaly detection.
\newblock In {\em Proc. IEEE Conf. Comp. Vis. Patt. Recogn.}, 2021.

\bibitem{scheirer2012toward}
Walter~J Scheirer, Anderson de Rezende~Rocha, Archana Sapkota, and Terrance~E
  Boult.
\newblock Toward open set recognition.
\newblock {\em {IEEE} Trans. Pattern Anal. Mach. Intell.}, 35(7):1757--1772,
  2012.

\bibitem{schlegl2019f}
Thomas Schlegl, Philipp Seeb{\"o}ck, Sebastian~M Waldstein, Georg Langs, and
  Ursula Schmidt-Erfurth.
\newblock f-anogan: Fast unsupervised anomaly detection with generative
  adversarial networks.
\newblock {\em Medical Image Analysis}, 54:30--44, 2019.

\bibitem{silvestre2019public}
Javier Silvestre-Blanes, Teresa Albero~Albero, Ignacio Miralles, Rub{\'e}n
  P{\'e}rez-Llorens, and Jorge Moreno.
\newblock A public fabric database for defect detection methods and results.
\newblock {\em Autex Research Journal}, 19(4):363--374, 2019.

\bibitem{sohn2020learning}
Kihyuk Sohn, Chun-Liang Li, Jinsung Yoon, Minho Jin, and Tomas Pfister.
\newblock Learning and evaluating representations for deep one-class
  classification.
\newblock In {\em Proc. Int. Conf. Learn. Representations}, 2021.

\bibitem{sultani2018real}
Waqas Sultani, Chen Chen, and Mubarak Shah.
\newblock Real-world anomaly detection in surveillance videos.
\newblock In {\em Proc. IEEE Conf. Comp. Vis. Patt. Recogn.}, pages 6479--6488,
  2018.

\bibitem{Tabernik2019JIM}
Domen Tabernik, Samo {\v{S}}ela, Jure Skvar{\v{c}}, and Danijel Sko{\v{c}}aj.
\newblock {Segmentation-Based Deep-Learning Approach for Surface-Defect
  Detection}.
\newblock {\em Journal of Intelligent Manufacturing}, May 2019.

\bibitem{tack2020csi}
Jihoon Tack, Sangwoo Mo, Jongheon Jeong, and Jinwoo Shin.
\newblock Csi: Novelty detection via contrastive learning on distributionally
  shifted instances.
\newblock {\em Proc. Advances in Neural Inf. Process. Syst.}, 33:11839--11852,
  2020.

\bibitem{taleb2007black}
Nassim~Nicholas Taleb.
\newblock {\em The black swan: The impact of the highly improbable}, volume~2.
\newblock Random house, 2007.

\bibitem{tian2021pixel}
Yu Tian, Yuyuan Liu, Guansong Pang, Fengbei Liu, Yuanhong Chen, and Gustavo
  Carneiro.
\newblock Pixel-wise energy-biased abstention learning for anomaly segmentation
  on complex urban driving scenes.
\newblock {\em arXiv: Comp. Res. Repository}, 2021.

\bibitem{tian2021constrained}
Yu Tian, Guansong Pang, Fengbei Liu, Yuanhong Chen, Seon~Ho Shin, Johan~W
  Verjans, Rajvinder Singh, and Gustavo Carneiro.
\newblock Constrained contrastive distribution learning for unsupervised
  anomaly detection and localisation in medical images.
\newblock In {\em Proc. Int. Conf. Medical Image Computing and Computer
  Assisted Intervention}, pages 128--140. Springer, 2021.

\bibitem{venkataramanan2020attention}
Shashanka Venkataramanan, Kuan-Chuan Peng, Rajat~Vikram Singh, and Abhijit
  Mahalanobis.
\newblock Attention guided anomaly localization in images.
\newblock In {\em Proc. Eur. Conf. Comp. Vis.}, pages 485--503. Springer, 2020.

\bibitem{wang2016s}
Peng Wang, Lingqiao Liu, Chunhua Shen, Zi Huang, Anton van~den Hengel, and
  Heng~Tao Shen.
\newblock What's wrong with that object? identifying images of unusual objects
  by modelling the detection score distribution.
\newblock In {\em Proc. IEEE Conf. Comp. Vis. Patt. Recogn.}, pages 1573--1581,
  2016.

\bibitem{wang2021glancing}
Shenzhi Wang, Liwei Wu, Lei Cui, and Yujun Shen.
\newblock Glancing at the patch: Anomaly localization with global and local
  feature comparison.
\newblock In {\em Proc. IEEE Conf. Comp. Vis. Patt. Recogn.}, pages 254--263,
  2021.

\bibitem{wang2019effective}
Siqi Wang, Yijie Zeng, Xinwang Liu, En Zhu, Jianping Yin, Chuanfu Xu, and
  Marius Kloft.
\newblock Effective end-to-end unsupervised outlier detection via inlier
  priority of discriminative network.
\newblock In {\em Proc. Advances in Neural Inf. Process. Syst.}, pages
  5960--5973, 2019.

\bibitem{wang2018revisiting}
Xinggang Wang, Yongluan Yan, Peng Tang, Xiang Bai, and Wenyu Liu.
\newblock Revisiting multiple instance neural networks.
\newblock {\em Pattern Recognition}, 74:15--24, 2018.

\bibitem{wieler2007weakly}
M Wieler and T Hahn.
\newblock Weakly supervised learning for industrial optical inspection.
\newblock In {\em DAGM Symposium}, 2007.

\bibitem{xiao2017_online}
Han Xiao, Kashif Rasul, and Roland Vollgraf.
\newblock Fashion-mnist: a novel image dataset for benchmarking machine
  learning algorithms, 2017.

\bibitem{yi2020patch}
Jihun Yi and Sungroh Yoon.
\newblock Patch {SVDD}: Patch-level svdd for anomaly detection and
  segmentation.
\newblock In {\em Proc. Asian Conf. Comp. Vis.}, 2020.

\bibitem{Yue_2021_CVPR}
Zhongqi Yue, Tan Wang, Qianru Sun, Xian-Sheng Hua, and Hanwang Zhang.
\newblock Counterfactual zero-shot and open-set visual recognition.
\newblock In {\em Proc. IEEE Conf. Comp. Vis. Patt. Recogn.}, pages
  15404--15414, June 2021.

\bibitem{yun2019cutmix}
Sangdoo Yun, Dongyoon Han, Seong~Joon Oh, Sanghyuk Chun, Junsuk Choe, and
  Youngjoon Yoo.
\newblock Cutmix: Regularization strategy to train strong classifiers with
  localizable features.
\newblock In {\em Proc. IEEE Int. Conf. Comp. Vis.}, pages 6023--6032, 2019.

\bibitem{Zaeemzadeh_2021_CVPR}
Alireza Zaeemzadeh, Niccolo Bisagno, Zeno Sambugaro, Nicola Conci, Nazanin
  Rahnavard, and Mubarak Shah.
\newblock Out-of-distribution detection using union of 1-dimensional subspaces.
\newblock In {\em Proc. IEEE Conf. Comp. Vis. Patt. Recogn.}, pages 9452--9461,
  June 2021.

\bibitem{zaheer2020old}
Muhammad~Zaigham Zaheer, Jin-ha Lee, Marcella Astrid, and Seung-Ik Lee.
\newblock Old is gold: Redefining the adversarially learned one-class
  classifier training paradigm.
\newblock In {\em Proc. IEEE Conf. Comp. Vis. Patt. Recogn.}, pages
  14183--14193, 2020.

\bibitem{zaheer2020claws}
Muhammad~Zaigham Zaheer, Arif Mahmood, Marcella Astrid, and Seung-Ik Lee.
\newblock Claws: Clustering assisted weakly supervised learning with normalcy
  suppression for anomalous event detection.
\newblock In {\em Proc. Eur. Conf. Comp. Vis.}, pages 358--376. Springer, 2020.

\bibitem{zhang2020viral}
Jianpeng Zhang, Yutong Xie, Guansong Pang, Zhibin Liao, Johan Verjans, Wenxing
  Li, Zongji Sun, Jian He, Yi Li, Chunhua Shen, and Yong Xia.
\newblock Viral pneumonia screening on chest x-rays using confidence-aware
  anomaly detection.
\newblock {\em IEEE T. Medical Imaging}, 40(3):879--890, 2021.

\bibitem{zhou2017anomaly}
Chong Zhou and Randy~C Paffenroth.
\newblock Anomaly detection with robust deep autoencoders.
\newblock In {\em Proc. {ACM SIGKDD} Int. Conf. Knowledge Discovery \& Data
  Mining}, pages 665--674, 2017.

\bibitem{zhou2021learning}
Da-Wei Zhou, Han-Jia Ye, and De-Chuan Zhan.
\newblock Learning placeholders for open-set recognition.
\newblock In {\em Proc. IEEE Conf. Comp. Vis. Patt. Recogn.}, pages 4401--4410,
  2021.

\bibitem{zhou2020encoding}
Kang Zhou, Yuting Xiao, Jianlong Yang, Jun Cheng, Wen Liu, Weixin Luo, Zaiwang
  Gu, Jiang Liu, and Shenghua Gao.
\newblock Encoding structure-texture relation with p-net for anomaly detection
  in retinal images.
\newblock In {\em Proc. Eur. Conf. Comp. Vis.}, pages 360--377. Springer, 2020.

\end{thebibliography}
}
\clearpage

\appendix
\section{Dataset Details}\label{app:dataset}

\subsection{Key Statistics of Datasets}
Tab. \ref{tab:stat} summarizes the key statistics of these datasets. Below we introduce each dataset in detail. The normal samples in MVTec AD are split into training and test sets following the original settings. In other datasets, the normal samples are randomly split into training and test sets by a ratio of $3/1$.

\textbf{MVTec AD} \cite{Bergmann_2019_CVPR} is a popular defect inspection benchmark that has 15 different classes, with each anomaly class containing one to several subclasses. In total the dataset contains 73 defective classes of fine-grained anomaly classes at the texture- or object-level.

\textbf{AITEX} \cite{silvestre2019public} is a fabrics defect inspection dataset that has 12 defect classes, with pixel-level defect annotation. We crop the original $4096\times256$ image to several $256\times256$ patch image and relabel each patch by pixel-wise annotation.

\textbf{SDD} \cite{Tabernik2019JIM} is a defect inspection dataset images of defective production items with pixel-level defect annotation. We vertically and equally divide the original $500\times1250$ image into three segment images and relabel each image by pixel-wise annotation.

\textbf{ELPV} \cite{deitsch2019elpv} is a solar cells defect inspection dataset in electroluminescence imagery. It contains two defect classes depending on solar cells: mono- and poly-crystalline.

\textbf{Optical} \cite{wieler2007weakly} is a synthetic dataset for defect detection on industrial optical inspection. The artificially generated data is similar to real-world tasks.

\textbf{Mastcam} \cite{kerner2020comparison} is a novelty detection dataset constructed from geological image taken by a multispectral imaging system installed in Mars exploration rovers. It contains typical images and images of 11 novel geologic classes. Images including shorter wavelength (color) channel and longer wavelengths (grayscale) channel and we focus on shorter wavelength channel in this work.

\textbf{BrainMRI} \cite{salehi2021multiresolution} is a brain tumor detection dataset obtained by magnetic resonance imaging (MRI) of the brain.

\textbf{HeadCT} \cite{salehi2021multiresolution} is a brain hemorrhage detection dataset obtained by CT scan of head.

\textbf{Hyper-Kvasir} \cite{borgli2020hyperkvasir} is a large-scale open gastrointestinal dataset collected during real gastro- and colonoscopy procedures. It contains four main categories and 23 subcategories of gastro- and colonoscopy images. This work focuses on gastroscopy images with the anatomical landmark category as the normal samples and the pathological category as the anomalies.
\begin{table}[t]
  \centering
  \caption{Key Statistics of Image Datasets. The first 15 datasets compose the MVTec AD dataset.}
  \scalebox{0.69}{
    \begin{tabular}{l@{}|ccc|cc}
    \hline
    \multirow{2}{*}{  \textbf{Dataset} }   & \textbf{Original Training} & \multicolumn{2}{c|}{\textbf{Original Test}} & \multicolumn{2}{c}{\textbf{Anomaly Data}}  \\
          \cline{2-6}
          & \textbf{Normal} & \textbf{Normal} & \textbf{Anomaly} & \textbf{\# Classes} & \textbf{Type} \\
          \hline
    Carpet & 280   & 28    & 89    & 5 & Texture\\
    Grid  & 264   & 21    & 57    & 5  & Texture\\
    Leather & 245   & 32    & 92    & 5 & Texture \\
    Tile  & 230   & 33    & 84    & 5  & Texture \\
    Wood  & 247   & 19    & 60    & 5  & Texture \\
    Bottle & 209   & 20    & 63    & 3  & Object \\
    Capsule & 219   & 23    & 109   & 5 & Object \\
    Pill  & 267   & 26    & 141   & 7 & Object \\
    Transistor & 213   & 60    & 40    & 4 & Object \\
    Zipper & 240   & 32    & 119   & 7 & Object \\
    Cable & 224   & 58    & 92    & 8& Object  \\
    Hazelnut & 391   & 40    & 70    & 4 & Object \\
    Metal\_nut & 220   & 22    & 93    & 4 & Object \\
    Screw & 320   & 41    & 119   & 5 & Object \\
    Toothbrush & 60    & 12    & 30    & 1& Object\\
    \hline
    \textbf{MVTec AD} & 3,629 &	467	& 1,258 & 73 & - \\\hline
    \textbf{AITEX} & 1,692  & 564   & 183   & 12& Texture  \\
    \textbf{SDD }  & 594   & 286   & 54    & 1 & Texture \\
    \textbf{ELPV}  & 1,131  & 377   & 715   & 2 & Texture \\
    \textbf{Optical} & 10,500 & 3,500  & 2,100  & 1& Object  \\
    \textbf{Mastcam}& 9,302  & 426   & 451   & 11 & Object \\
    \textbf{BrainMRI} & 73    & 25    & 155   & 1 & Medical \\
   \textbf{HeadCT} & 75    & 25    & 100   & 1 & Medical \\
    \textbf{Hyper-Kvasir} & 2,021  & 674   & 757   & 4& Medical \\
    \hline
    \end{tabular}%
    }
  \label{tab:stat}%
\end{table}%
\begin{table}[t]
  \centering
  \caption{Download Link of Image Datasets. }
  \scalebox{0.69}{
    \begin{tabular}{l@{}|l}
    \hline
     \textbf{Dataset}   & \textbf{Link} \\
          \hline
    \textbf{MVTec AD} & \url{https://tinyurl.com/mvtecad}\\
    \textbf{AITEX}  &\url{https://tinyurl.com/aitex-defect}\\
    \textbf{SDD }   & \url{https://tinyurl.com/KolektorSDD}\\
    \textbf{ELPV}   & \url{https://tinyurl.com/elpv-crack}\\
    \textbf{Optical} & \url{https://tinyurl.com/optical-defect}\\
    \textbf{Mastcam} & \url{https://tinyurl.com/mastcam}\\
    \textbf{BrainMRI}  & \url{https://tinyurl.com/brainMRI-tumor}\\
   \textbf{HeadCT} &  \url{https://tinyurl.com/headCT-tumor}\\
    \textbf{Hyper-Kvasir} &  \url{https://tinyurl.com/hyper-kvasir}\\
    \hline
    \end{tabular}%
    }
  \label{tab:link}%
\end{table}%

\begin{figure*}[h!]
  \centering
    \includegraphics[width=0.92\textwidth]{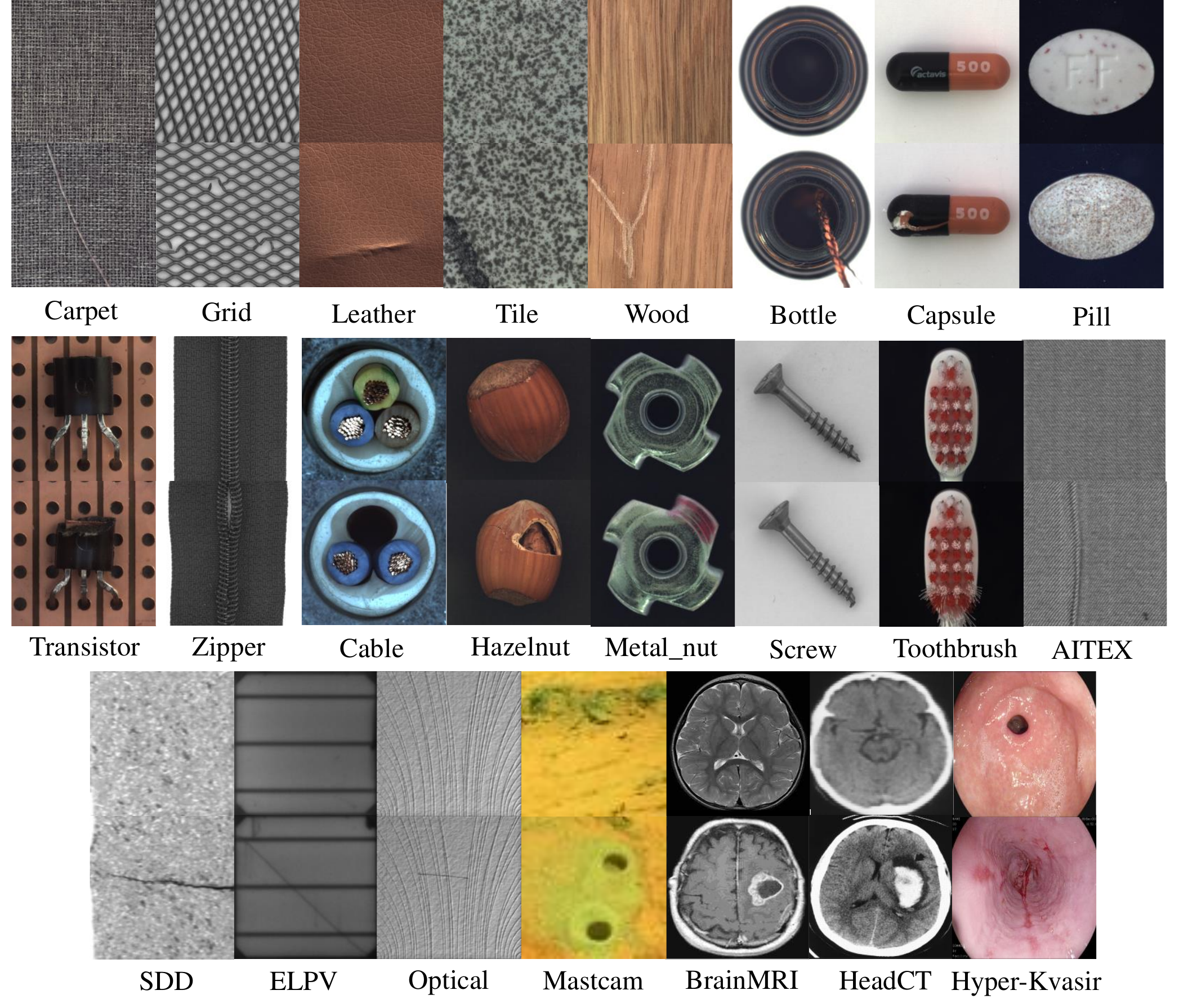}
  \caption{Examples of normal and anomalous images for each dataset. For each group of examples, the images on the top are normal, while the bottom ones are anomalous.}
  \label{fig:dataset}
\end{figure*}

To provide some intuitive understanding of what the anomalies and normal samples look like, we present some examples of normal and anomalous images for each dataset in Fig. \ref{fig:dataset}.

\subsection{Dataset Split}
We have two experiment protocols, including general and hard settings.
For the general setting, the few labeled anomaly samples are randomly drawn from all possible anomaly classes in the test set per dataset. These sampled anomalies are then removed from the test data. For the hard setting, the anomaly example sampling is limited to be drawn from one single anomaly class only, and all anomaly samples in this anomaly class are removed from the test set to ensure that the test set contains only unseen anomaly classes. As labeled anomalies are difficult to obtain due to their rareness and unknowingness, in both settings we use only very limited labeled anomalies, i.e., with the number of the given anomaly examples respectively fixed to one and ten.

Additionally, to have a cross analysis of the results in the one-shot and ten-shot scenarios, the one anomaly example in the one-shot scenarios is randomly sampled from the ten sampled anomaly examples in the ten-shot scenarios, and they are all evaluated on exactly the same test data -- the test data used in ten-shot scenarios. That is, the only difference between the one-shot and ten-shot scenarios is on the training anomaly examples.

\section{Implementation Details}
In this section, we describe the implementation details of DRA and its competing methods.
\subsection{Implementation of DRA}\label{app:implementation}
All input images are first resized to 448x448 or 224x224 according to the original resolution. We then use the ImageNet pre-trained ResNet-18 for the feature extraction network, which extracts a 512-dimensional feature map for an input image. This feature map is then fed to the subsequent abnormality/normality learning heads to learn disentangled abnormalities. The Patch-wise classifier in Plain Feature Learning adopted in the seen and pseudo abnormality learning heads is implemented by a 1x1 convolutional layer that yields the anomaly score of each vector in the feature map. The normality learning head utilizes a two-layer fully connected layer as the classifier, which first reduces a 512-dimensional feature vector to 256-dimensions and then yields anomaly scores. In the training phase, each input image is routed to different heads based on the labels, and each head computes the loss independently. All its heads are jointly trained using 30 epochs, with 20 iterations per epoch and a batch size of 48. Adam is used for the parameter optimization using an initial learning rate $10^{-3}$ with a weight decay of $10^{-2}$.

In the inference phase, each head yields anomaly scores for the target image simultaneously. Since all heads are optimized by the same loss function, the anomaly scores generated by each head have the same semantic, and so we calculate the sum of all the anomaly scores (and a negated normal score) as the final anomaly score. In addition, to solve the multi-scale problem, we use an image pyramid module with two-layer image pyramid, which obtains anomaly scores at different scales by inputting original images of various sizes, and calculates the mean value as the final anomaly score. 

For the reference sets in Latent Residual Abnormality Learning, we found that mixing some generated pseudo-anomaly samples into normal samples can further improve performance. We speculate that adding pseudo-anomaly samples can get more challenging residual samples to help the network adapt to extreme cases. Therefore, we use the dataset mixed with normal and pseudo-abnormal samples as the reference set in the final implementation.

\subsubsection{Loss Function}
DRA can be optimized using different anomaly score loss functions. We use deviation loss \cite{pang2021explainable} in our final implementation to optimize DRA, because it is generally more effective and stable than other popular loss functions, as shown in the experimental results in Section \ref{subsec:loss}. Particularly, a deviation loss optimizes the anomaly scoring network by a Gaussian prior score, with the deviation specified as a Z-score:
\begin{equation}\label{eqn:deviation}
    \mathit{dev}(\mathbf{x}_{i};\Theta) = \frac{g(f (\mathbf{x};\Theta_f);\Theta_{g}) - \mu_{\mathcal{R}}}{\sigma_{\mathcal{R}}},
\end{equation}
where $\mu_{\mathcal{R}}$ and $\sigma_{\mathcal{R}}$ is the mean and standard deviation of the prior-based anomaly score set drawn from ${\cal N} (\mu, \sigma^{2})$. The deviation loss is specified using the contrastive loss \cite{hadsell2006contrastloss} with the deviation plugged into:
\begin{multline}\label{eqn:loss}
    \ell\big(\mathbf{x}_{i}, \mu_{\mathcal{R}}, \sigma_{\mathcal{R}};\Theta \big) = (1-y_{i})|\mathit{dev}(\mathbf{x}_{i};\Theta)|\\ + y_{i} \max\big(0, a - \mathit{dev}(\mathbf{x}_{i};\Theta)\big),
\end{multline}
where $y = 1$ indicate an anomaly and $y = 0$ indicate a normal sample,
and $a$ is equivalent to a Z-Score confidence interval parameter.

\subsection{Implementation of Competing Methods}
In the main text, we present five recent and closely related state-of-the-art (SOTA) competing methods. Here we introduce two additional competing methods. Following is the detailed description and implementation details of these seven methods:

\noindent\textbf{KDAD} \cite{salehi2021multiresolution} is an unsupervised deep anomaly detector based on multi-resolution knowledge distillation. We experiment with the code provided by its authors\footnote{\url{https://github.com/rohban-lab/Knowledge\_Distillation\_AD}} and report the results. Since KDAD is unsupervised, it is trained with normal data only, but it is evaluated on exactly the same test data as DRA.

\noindent\textbf{DevNet} \cite{pang2019deep,pang2021explainable} is a supervised deep anomaly detector based on a prior-based deviation. The results we report are based on the implementation provided by its authors\footnote{\url{https://github.com/choubo/deviation-network-image}}.

\noindent\textbf{FLOS} \cite{lin2017focalloss} is a deep imbalanced classifier that learns a binary classification model using the class-imbalance-sensitive loss -- focal loss. The implementation of FLOS is also taken from \cite{pang2021explainable}, which replaces the loss function of DevNet with the focal loss.

\noindent\textbf{SAOE} is a deep out-of-distribution detector that utilizes pseudo anomalies from both data augmentation-based and outlier exposure-based methods. Motivated by the success of using pseudo anomalies to improve anomaly detection in recent studies \cite{tack2020csi,li2021cutpaste}, SAOE is implemented by learning both seen and pseudo abnormalities through a multi-class (\ie, normal class, seen anomaly class, and pseudo anomaly class) classification head using the plain feature learning method as in DRA. In addition to this multi-class classification, the outlier exposure module \cite{hendrycks2018deep} in SAOE is implemented according to its authors\footnote{\url{https://github.com/hendrycks/outlier-exposure}}, in which the MVTec AD \cite{Bergmann_2019_CVPR} or LAG \cite{Li_2019_CVPR} dataset is used as external data. In all our experiments we removed the related data from the outlier data that has any overlapping with the target data to avoid data leakage.

\noindent\textbf{MLEP} \cite{liu2019margin} is a deep open set anomaly detector based on margin learning embedded prediction. The original MLEP\footnote{\url{https://github.com/svip-lab/MLEP}} is designed for open set video anomaly detection, and we adapt it to image tasks by modifying the backbone network and training settings to be consistent with DRA.

\noindent\textbf{Deep SAD} \cite{ruff2019deep} is a supervised deep anomaly detector that extends Deep SVDD \cite{ruff2018deep} by using a few labeled anomalies and normal samples to learn more compact one-class descriptors. Particularly, it adds a new marginal constraint to the original Deep SVDD that enforces a large margin between labeled anomalies and the one-class center in latent space. The implementation of DeepSAD is taken from the original authors\footnote{\url{https://github.com/lukasruff/Deep-SAD-PyTorch}}.

\noindent\textbf{MINNS} \cite{wang2018revisiting} is a deep multiple instance classification model, which is implemented based on \cite{pang2021explainable}.

\begin{table}[bt]
  \centering
  \caption{AUC results (mean±std) of DRA and two additional competing methods under the general setting. All methods are trained using ten random anomaly examples, with the best results are \textbf{highlighted}.
}
  \vspace{-0.2cm}
  \scalebox{0.77}{
    \begin{tabular}{l@{}|c|ccc}
    \hline
    \textbf{Dataset} & $|\mathcal{C}|$ & \multicolumn{1}{c}{\textbf{DeepSAD}} & \multicolumn{1}{c}{\textbf{MINNS}} & \multicolumn{1}{c}{\textbf{DRA (Ours)}} \\\hline
    Carpet & 5 & 0.791\footnotesize{±0.011}& 0.876\footnotesize{±0.015}& \textbf{0.940}\footnotesize{±0.027} \\
    
    Grid & 5 & 0.854\footnotesize{±0.028}& 0.983\footnotesize{±0.016}& \textbf{0.987}\footnotesize{±0.009} \\
    
    Leather & 5 & 0.833\footnotesize{±0.014}& 0.993\footnotesize{±0.007}& \textbf{1.000}\footnotesize{±0.000}\\
    
    Tile & 5 & 0.888\footnotesize{±0.010}& 0.980\footnotesize{±0.003}& \textbf{0.994}\footnotesize{±0.006}\\
    
    Wood & 5 & 0.781\footnotesize{±0.001}& \textbf{0.998}\footnotesize{±0.004}& \textbf{0.998}\footnotesize{±0.001} \\

    Bottle & 3 & 0.913\footnotesize{±0.002}& 0.995\footnotesize{±0.007}& \textbf{1.000}\footnotesize{±0.000} \\
    
    Capsule & 5 & 0.476\footnotesize{±0.022}& 0.905\footnotesize{±0.013}& \textbf{0.935}\footnotesize{±0.022} \\
    
    Pill & 7 & 0.875\footnotesize{±0.063}& \textbf{0.913}\footnotesize{±0.021}& 0.904\footnotesize{±0.024}\\
    
    Transistor & 4 & 0.868\footnotesize{±0.006}& 0.889\footnotesize{±0.032}& \textbf{0.915}\footnotesize{±0.025}\\
    
    Zipper & 7 & 0.974\footnotesize{±0.005}& 0.981\footnotesize{±0.011}& \textbf{1.000}\footnotesize{±0.000} \\
    
    Cable &8 & 0.696\footnotesize{±0.016}& 0.842\footnotesize{±0.012}& \textbf{0.909}\footnotesize{±0.011} \\
    
    Hazelnut &4 & \textbf{1.000}\footnotesize{±0.000}& \textbf{1.000}\footnotesize{±0.000}& \textbf{1.000}\footnotesize{±0.000} \\
    
    Metal\_nut & 4 & 0.860\footnotesize{±0.053}& 0.984\footnotesize{±0.002}& \textbf{0.997}\footnotesize{±0.002} \\
    
    Screw & 5 & 0.774\footnotesize{±0.081}& 0.932\footnotesize{±0.035}& \textbf{0.977}\footnotesize{±0.009} \\
    
    Toothbrush & 1 & \textbf{0.885}\footnotesize{±0.063}& 0.810\footnotesize{±0.086}& 0.826\footnotesize{±0.021} \\
    
    \hline
    \textbf{MVTec AD} & - & 0.830\footnotesize{±0.009}& 0.939\footnotesize{±0.011}& \textbf{0.959}\footnotesize{±0.003} \\
    
    \textbf{AITEX } &12 & 0.686\footnotesize{±0.028}& 0.813\footnotesize{±0.030}& \textbf{0.893}\footnotesize{±0.017} \\
    
    \textbf{SDD}& 1 & 0.963\footnotesize{±0.005}& 0.961\footnotesize{±0.016}& \textbf{0.991}\footnotesize{±0.005} \\
    
    \textbf{ELPV} & 2 & 0.722\footnotesize{±0.053}& 0.788\footnotesize{±0.028}& \textbf{0.845}\footnotesize{±0.013} \\
    
   \textbf{Optical}& 1 & 0.558\footnotesize{±0.012}& 0.774\footnotesize{±0.047}& \textbf{0.965}\footnotesize{±0.006}\\
   
   \textbf{Mastcam}& 11 & 0.707\footnotesize{±0.011}& 0.803\footnotesize{±0.031}& \textbf{0.848}\footnotesize{±0.008}\\
   
    \textbf{BrainMRI}& 1 & 0.850\footnotesize{±0.016}& 0.943\footnotesize{±0.031}& \textbf{0.970}\footnotesize{±0.003}\\
    
   \textbf{HeadCT}& 1 & 0.928\footnotesize{±0.005}& \textbf{0.984}\footnotesize{±0.010}& 0.972\footnotesize{±0.002}\\
   
   \textbf{Hyper-Kvasir$\ $}& 4 & 0.719\footnotesize{±0.032}& 0.647\footnotesize{±0.051}& \textbf{0.834}\footnotesize{±0.004} \\
   \hline
    \end{tabular}
    }
  \label{tab:randomanomalies}%
  \vspace{-0.3cm}
\end{table}%
\section{Additional Empirical Results}
\subsection{Additional Comparison Results}
\noindent\textbf{General Setting.} We report the results of DRA and two additional competing methods under general setting in Tab \ref{tab:randomanomalies}. Our method achieves the best AUC performance in eight of the nine datasets and the close-to-best AUC performance in the another dataset. In the eight best-performing datasets, our method improves AUC by 2\% to 19.1\% over the best competing method.

\noindent\textbf{Hard Setting.} Tab. \ref{tab:hard} shows the results of DRA and two additional competing methods under the hard setting. Our method performs best on most of the data subsets and achieves the best AUC performance on five of the six datasets at the dataset level. 
Our method improves from 9.2\% to 24.4\% over the suboptimal method in the other five datasets.

The experimental results in both settings show the superiority of our method compared to Deep SAD and MINNS.

\begin{table}[tb]
  \centering
  \caption{AUC results of DRA and two additional competing methods under the hard setting, where models are trained with one known anomaly class and tested to detect the rest of all other anomaly classes. Each data subset is named by the known anomaly class.}
  \vspace{-0.2cm}
  \scalebox{0.85}{
    \begin{tabular}{p{0.2cm}p{1.92cm}|ccc}
    \hline
    \multicolumn{2}{l}{\textbf{Module}}   & \multicolumn{1}{|c}{\textbf{DeepSAD}} & \multicolumn{1}{c}{\textbf{MINNS}} & \multicolumn{1}{c}{\textbf{DRA (Ours)}} \\ 
    \hline
    \multirow{6}{*}{\rotatebox{90}{\textbf{Carpet}}} & Color & 0.736\footnotesize{±0.007}& 0.767\footnotesize{±0.011}& \textbf{0.886}\footnotesize{±0.042} \\
          & Cut & 0.612\footnotesize{±0.034}& 0.694\footnotesize{±0.068}& \textbf{0.922}\footnotesize{±0.038}\\
          & Hole & 0.576\footnotesize{±0.036}& 0.766\footnotesize{±0.007}& \textbf{0.947}\footnotesize{±0.016}\\
          & Metal & 0.732\footnotesize{±0.042}& 0.789\footnotesize{±0.097}& \textbf{0.933}\footnotesize{±0.022}\\
          & Thread & 0.979\footnotesize{±0.000}& 0.982\footnotesize{±0.008}& \textbf{0.989}\footnotesize{±0.004}\\
          \cline{2-5}
          & \textbf{Mean} & 0.727\footnotesize{±0.011}& 0.800\footnotesize{±0.022}& \textbf{0.935}\footnotesize{±0.013} \\
    \hline
    \multirow{5}[0]{*}{\rotatebox{90}{\textbf{Metal\_nut}}} & Bent & 0.821\footnotesize{±0.023}& 0.868\footnotesize{±0.033}& \textbf{0.990}\footnotesize{±0.003}\\
          & Color & 0.707\footnotesize{±0.028}& \textbf{0.985}\footnotesize{±0.018}& 0.967\footnotesize{±0.011} \\
          & Flip & 0.602\footnotesize{±0.020}& \textbf{1.000}\footnotesize{±0.000}& 0.913\footnotesize{±0.021} \\
          & Scratch & 0.654\footnotesize{±0.004}& \textbf{0.978}\footnotesize{±0.000}& 0.911\footnotesize{±0.034} \\
          \cline{2-5}
          & \textbf{Mean} & 0.696\footnotesize{±0.012}& \textbf{0.958}\footnotesize{±0.008}& 0.945\footnotesize{±0.017}  \\
    \hline
    \multirow{7}[0]{*}{\rotatebox{90}{\textbf{AITEX}}} & Broken\_end & 0.442\footnotesize{±0.029}& \textbf{0.708}\footnotesize{±0.103}& 0.693\footnotesize{±0.099} \\
          & Broken\_pick & 0.614\footnotesize{±0.039}& 0.565\footnotesize{±0.018}& \textbf{0.760}\footnotesize{±0.037} \\
          & Cut\_selvage & 0.523\footnotesize{±0.032}& 0.734\footnotesize{±0.012}& \textbf{0.777}\footnotesize{±0.036}\\
          & Fuzzyball & 0.518\footnotesize{±0.023}& 0.534\footnotesize{±0.058}& \textbf{0.701}\footnotesize{±0.093}\\
          & Nep & 0.733\footnotesize{±0.017}& 0.707\footnotesize{±0.059}& \textbf{0.750}\footnotesize{±0.038} \\
          & Weft\_crack & 0.510\footnotesize{±0.058}& 0.544\footnotesize{±0.183}& \textbf{0.717}\footnotesize{±0.072}\\
          \cline{2-5}
          & \textbf{Mean} & 0.557\footnotesize{±0.014}& 0.632\footnotesize{±0.023}& \textbf{0.733}\footnotesize{±0.009} \\
    \hline
    \multirow{3}[0]{*}{\rotatebox{90}{\textbf{ELPV}}} & Mono & 0.554\footnotesize{±0.063}& 0.557\footnotesize{±0.010}& \textbf{0.731}\footnotesize{±0.021}\\
          & Poly & 0.621\footnotesize{±0.006}& 0.770\footnotesize{±0.032}& \textbf{0.800}\footnotesize{±0.064} \\
          \cline{2-5}
          & \textbf{Mean} & 0.588\footnotesize{±0.021}& 0.663\footnotesize{±0.015}& \textbf{0.766}\footnotesize{±0.029}\\
    \hline
    \multirow{10}[0]{*}{\rotatebox{90}{\textbf{Mastcam}}} 
          & Bedrock & 0.474\footnotesize{±0.038}& 0.419\footnotesize{±0.025}& \textbf{0.658}\footnotesize{±0.021} \\
          & Broken-rock & 0.497\footnotesize{±0.054}& \textbf{0.687}\footnotesize{±0.015}& 0.649\footnotesize{±0.047} \\
          & Drill-hole & 0.494\footnotesize{±0.013}& 0.651\footnotesize{±0.035}& \textbf{0.725}\footnotesize{±0.005}\\
          & Drt & 0.586\footnotesize{±0.012}& 0.705\footnotesize{±0.043}& \textbf{0.760}\footnotesize{±0.033} \\
          & Dump-pile & 0.565\footnotesize{±0.046}& 0.697\footnotesize{±0.022}& \textbf{0.748}\footnotesize{±0.066}\\
          & Float & 0.408\footnotesize{±0.022}& 0.635\footnotesize{±0.073}& \textbf{0.744}\footnotesize{±0.073} \\
          & Meteorite & 0.489\footnotesize{±0.010}& 0.551\footnotesize{±0.018}& \textbf{0.716}\footnotesize{±0.004} \\
          & Scuff & 0.502\footnotesize{±0.010}& 0.502\footnotesize{±0.040}& \textbf{0.636}\footnotesize{±0.086}\\
          & Veins & 0.542\footnotesize{±0.017}& 0.577\footnotesize{±0.013}& \textbf{0.620}\footnotesize{±0.036}\\
          \cline{2-5}
          & \textbf{Mean} & 0.506\footnotesize{±0.009}& 0.603\footnotesize{±0.016}& \textbf{0.695}\footnotesize{±0.004} \\
    \hline
    \multirow{5}{*}{\rotatebox{90}{\textbf{Hyper-Kvasir}}} & Barretts & 0.672\footnotesize{±0.017}& 0.679\footnotesize{±0.009}& \textbf{0.824}\footnotesize{±0.006}\\
          & 
          B.-short-seg & 0.666\footnotesize{±0.012}& 0.608\footnotesize{±0.064}& \textbf{0.835}\footnotesize{±0.021}\\
          & 
          Esophagitis-a & 0.619\footnotesize{±0.027}& 0.665\footnotesize{±0.045}& \textbf{0.881}\footnotesize{±0.035}\\
          & 
          E.-b-d & 0.564\footnotesize{±0.006}& 0.480\footnotesize{±0.043}& \textbf{0.837}\footnotesize{±0.009} \\
          \cline{2-5}
          & \textbf{Mean} & 0.630\footnotesize{±0.009}& 0.608\footnotesize{±0.014}& \textbf{0.844}\footnotesize{±0.009} \\
    \hline
    \end{tabular}%
    }
    \vspace{-0.2cm}
  \label{tab:hard}%
\end{table}%

\subsection{Sensitivity w.r.t. Loss Function}\label{subsec:loss}

\begin{figure}[t]
  \centering
    \includegraphics[width=0.42\textwidth]{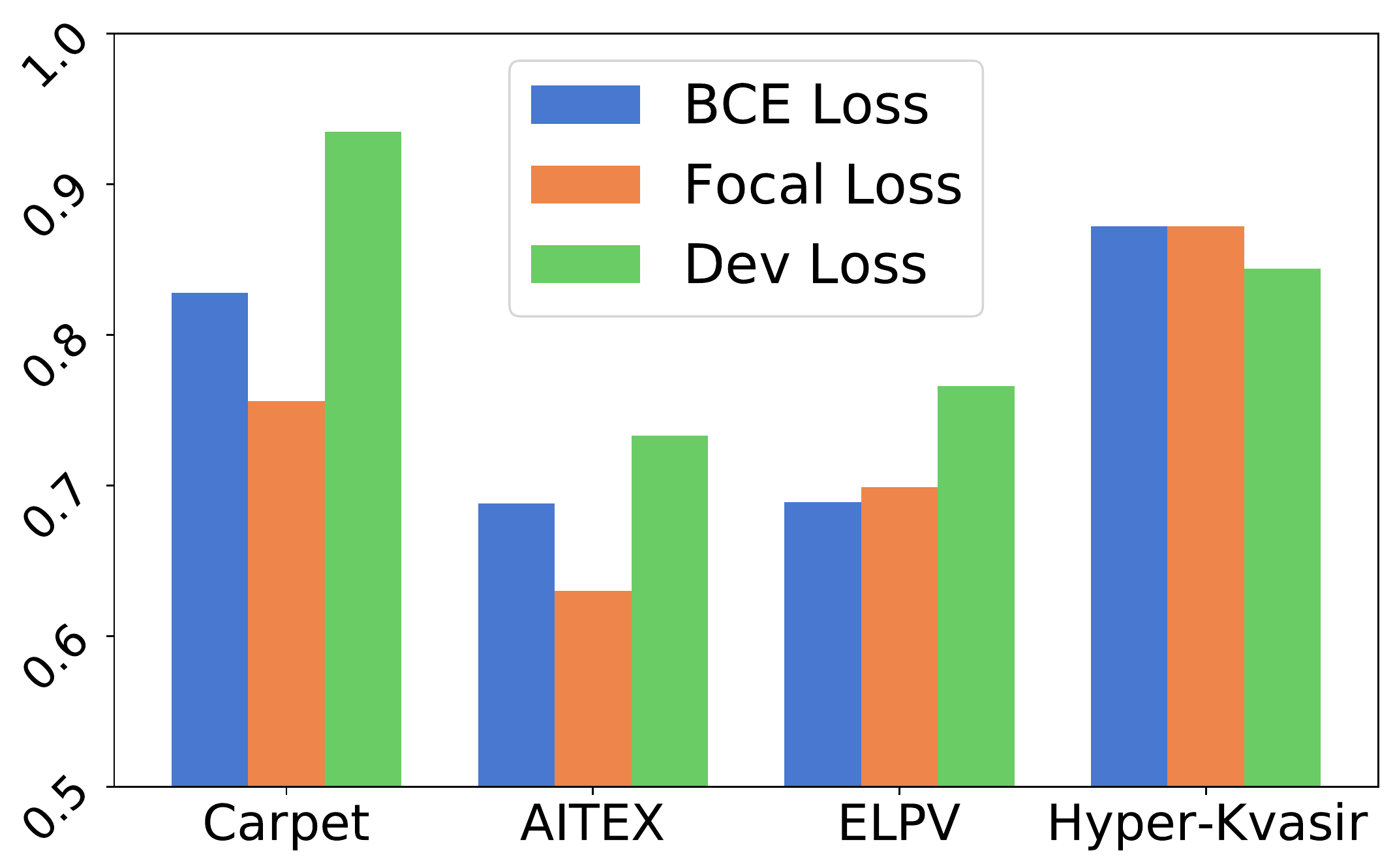}
  \caption{The AUC performance of our proposed method using different loss functions under the hard setting. We report the averaged results over all data subsets per dataset.}
  \label{fig:ana_loss}
\end{figure}

In our paper, we use the deviation loss \cite{pang2021explainable} in all our four heads by default. Here we vary the use of the loss function and analyze the impact of the loss function on the performance of anomaly detection. Any related binary classification loss functions may be used for training all the four heads of DRA.
We evaluate the applicability of two additional popular loss functions, including binary cross-entropy loss and focal loss, in addition to deviation loss. The results are reported in Fig. \ref{fig:ana_loss}, where all results are the averaged AUC of three independent runs of the experiments. 
In general, the deviation loss function, which is specifically designed for anomaly detection, has clear superiority on most cases. The two classification losses perform better on the medical dataset Hyper-Kvasir. Based on such empirical findings, the deviation loss function is generally recommended in DRA.

\subsection{Cross-domain Anomaly Detection}
An interesting extension area of open-set anomaly detection is cross-domain anomaly detection, aiming at training detection models on a source domain to detect anomalies on datasets from a target domain different from the source domain. To demonstrate potential of our method in such setting, we report cross-domain AD results of our model DRA on all five texture anomaly datasets in MVTec AD in Tab. \ref{tab: cross_domain}. DRA is trained on one of five datasets (source domain) and in fine-tuned with 10 epochs on the other four datasets (target domains) using normal samples only. The results show that the domain-adapted DRA significantly outperforms the SOTA unsupervised method KDAD that is directly trained on the target domain. This demonstrates some promising open-domain performance of DRA.
\begin{table}[ht]
\caption{AUC results of domain-adapted DRA and unsupervised method KDAD in texture datasets. The top row is the source domain and the left column is the target domain.}
\centering
\label{tab: cross_domain}
\scalebox{0.85}{
\begin{tabular}{c|cccccc} 
\hline
        & carpet   & grid    & leather   & tile & wood & KDAD    \\ 
\hline
\hline
carpet              & - & 0.833 & 0.921 & 0.930 & 0.917 & 0.774\\ 
\hline
grid                & 0.983 & - & 0.924 & 0.940 & 0.916 & 0.749\\ 
\hline
leather             & 0.988 & 0.998 & - & 0.994 & 1.000 & 0.948\\
\hline
tile                & 0.917 & 0.971 & 0.958 & - & 0.955 & 0.911\\
\hline
wood                & 0.993 & 0.985 & 0.972 & 0.948 & - & 0.940\\
\hline
\end{tabular}}
\label{tab:cross_domain}%
\end{table}

\section{Failure Cases}
Although DRA shows competitive results on most datasets, it still fails on individual datasets; the most notable is the toothbrush dataset. After in-depth research and analysis of the results, we believe the failure of the toothbrush dataset is mainly due to its small size of normal samples (60 normal samples, see Tab. \ref{tab:stat}). Due to the more complex architecture, DRA often requires a relatively larger set of normal training samples to learn the disentangled abnormalities, while simpler methods like FLOS and SAOE that perform mainly binary classification do not have this requirement and work better on this dataset. In practice, we need to pay attention to the available data size of the target task, and apply a lightweight network in DRA instead when facing small-scale tasks.

\end{document}